\documentclass[11pt]{article}

\usepackage[preprint]{acl}

\usepackage{times}
\usepackage{latexsym}

\usepackage[T1]{fontenc}

\usepackage[utf8]{inputenc}

\usepackage{microtype}

\usepackage{inconsolata}

\usepackage{graphicx}
\usepackage{booktabs}
\usepackage{multirow}
\usepackage{xcolor}
\usepackage{listings}
\usepackage{wrapfig}
\usepackage{tabularx}
\usepackage{array}
\usepackage{float}
\usepackage{graphicx}
\usepackage{fancyvrb}
\usepackage{subcaption}
\usepackage{afterpage}
\usepackage{placeins}
\usepackage{caption}
\usepackage{enumitem}
\usepackage[most]{tcolorbox}

\usepackage{pifont} 
\newcommand{\cmark}{\textcolor{green!70!black}{\ding{51}}}
\newcommand{\xmark}{\textcolor{red!70!black}{\ding{55}}}

\title{MI-CXR: A Benchmark for Longitudinal Reasoning over Multi-Interval Chest X-rays}

\author{
Sunghwan Steve Cho$^{1}$ \quad
Yunseok Han$^{2}$ \quad
Jaeyoung Do$^{1,2,\dagger}$ \\
AIDAS Laboratory, $^{1}$ECE \& $^{2}$IPAI, Seoul National University \\
\texttt{\{steve97, qicher, jaeyoung.do\}}@snu.ac.kr
}

\begin{document}

\maketitle
\footnotetext{$^{\dagger}$ Corresponding author}
\maketitle
\begin{abstract}
Longitudinal chest X-ray (CXR) interpretation requires reasoning over disease evolution across multiple patient visits, yet most existing medical VQA benchmarks focus on single images or short-horizon image pairs.
We introduce \textbf{MI-CXR}, a benchmark for standardized evaluation of \textbf{M}ulti-\textbf{I}nterval longitudinal reasoning over multi-visit \textbf{CXR} sequences, without requiring free-form report generation or additional clinical context.
MI-CXR comprises five-way multiple-choice questions over five-visit patient timelines and instantiates three complementary task families:
\emph{Temporal Event Localization}, \emph{Interval-wise Change Reasoning}, and \emph{Global Trajectory Summarization}, which assess clinically grounded visual reasoning over time.
Evaluating 14 state-of-the-art vision--language models (VLMs) shows low overall performance (29.3\% accuracy), only modestly above random guessing.
Using stage-wise diagnostic probing, we find that models often produce locally plausible interval descriptions but fail to enforce temporal constraints or compose evidence into globally consistent decisions over the full timeline.
These findings reveal key limitations of current VLMs and establish MI-CXR as a principled benchmark for longitudinal medical reasoning. The benchmark is available at \url{https://github.com/AIDASLab/MI-CXR}.
\end{abstract}

\section{Introduction}
\label{sec:intro}

\begin{figure*}[t]
\centering
\includegraphics[width=\linewidth]{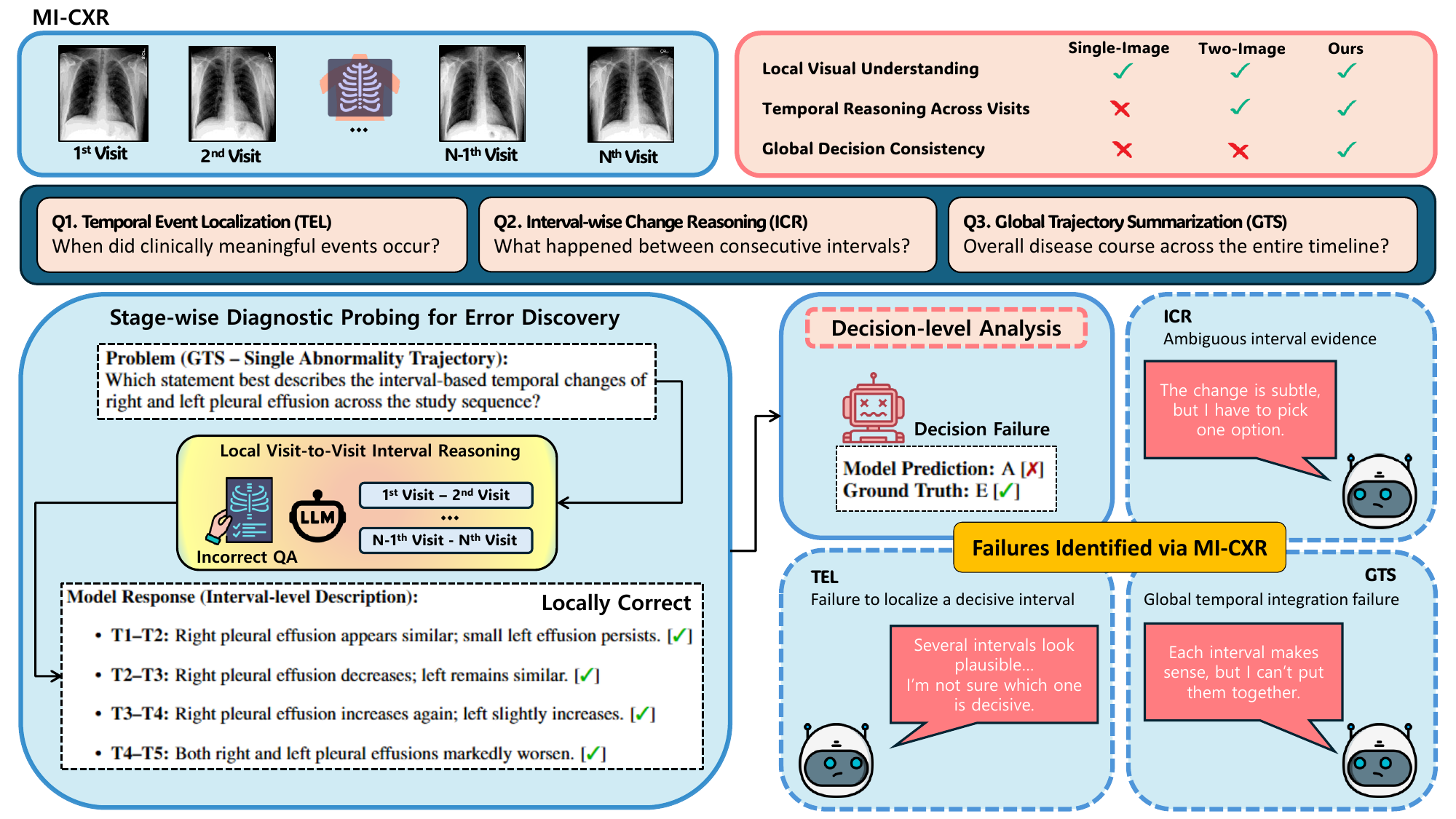} 
\caption{\textbf{Overview of longitudinal medical visual question answering and MI-CXR.}
Clinical image interpretation requires integrating evidence across multiple patient visits (top).
We formalize longitudinal medical VQA into three core reasoning capabilities—Temporal Event Localization (TEL),
Interval-wise Change Reasoning (ICR), and Global Trajectory Summarization (GTS)—and evaluate them over multi-visit CXR sequences using a diagnostic stage-wise decomposition (bottom).}
\label{fig:intro_fig_1}
\vspace{-0.7em}
\end{figure*}

Despite rapid progress in vision--language models (VLMs) for medical image understanding, most existing evaluations adopt simplified problem formulations that diverge from real clinical workflows~\cite{vqa-rad, mimic-diff-vqa, mmxu}.
In chest X-ray (CXR) interpretation, diagnostic reasoning rarely relies on isolated images; instead, clinicians routinely compare examinations acquired across multiple visits to assess disease onset, progression, response to treatment, and recurrence~\cite{olex2021temporal, acosta2022prior, jin2021predicting}.

However, current CXR medical benchmarks predominantly focus on restricted settings, such as single-image recognition~\cite{vqa-rad, pathvqa} or pairwise image comparison~\cite{mmxu, mimic-diff-vqa, temmed}.
While these formulations capture important sub-problems, they fail to support many clinically meaningful questions that are \emph{inherently longitudinal}, including when an abnormality first appears, whether it recurs after resolution, and how disease evolves over time~\cite{vantimmeren2025longitudinal}.


A key challenge is that longitudinal interpretation imposes constraints absent in single-image or pairwise settings.
Clinical decisions must remain \emph{globally consistent} across temporally ordered visits, resolving mutually exclusive event hypotheses and composing local changes into coherent trajectory-level conclusions~\cite{ACR_Diagnostic_CT, diagnostics12071735, White1994_PriorRadiographs, Zhang2023_DiagnosticErrorRadiology, mimiccxr}.
As a result, even when VLMs can describe local interval-level changes~\cite{lasateam2025lingshugeneralistfoundationmodel, biomed_gpt, sellergren2025medgemmatechnicalreport, pan2025medvlmr1incentivizingmedicalreasoning, li2023llavamedtraininglargelanguageandvision}, they may still fail at temporal diagnostic reasoning that requires structured decision-making over extended, dependent evidence. To address this mismatch, we formalize medical VQA as a problem of multi-interval longitudinal reasoning over multi-visit CXR sequences.
Rather than treating longitudinal understanding as a straightforward extension of pairwise comparison, we decompose it into three core reasoning capabilities that naturally arise in clinical workflows and jointly stress different aspects of temporal reasoning (Figure~\ref{fig:intro_fig_1}).

Specifically, \textbf{Temporal Event Localization (TEL)} requires identifying when clinically meaningful events—such as abnormality emergence, resolution, or recurrence—occur along the timeline, emphasizing decisive reasoning under temporal ordering and exclusivity constraints~\cite{xu2025bleedorigindynamicbleedingsource, Mann2025}.
\textbf{Interval-wise Change Reasoning (ICR)} focuses on interpreting visual changes between consecutive visits, isolating local interval-level perception that underlies longitudinal interpretation~\cite{Hoang2016}.
\textbf{Global Trajectory Summarization (GTS)} further requires integrating evidence across all visits to characterize the overall disease course, making decisions sensitive to cumulative context and error propagation~\cite{Holste2024, VANTIMMEREN2025103610}.

Based on this formulation, we introduce \textbf{MI-CXR}, a benchmark for evaluating \textbf{M}ulti-\textbf{I}nterval longitudinal reasoning over multi-visit \textbf{CXR} sequences.
The benchmark consists of curated multi-visit patient timelines paired with questions that explicitly target the above reasoning capabilities (i.e., TEL, ICR, and GTS).
Crucially, each question is constructed such that correct answers require aggregating information across multiple visits, rather than relying on cues from any single image or isolated image pair.
This enables a principled and fine-grained assessment of whether models can reason over extended visual evidence in a clinically meaningful manner.
Our evaluation under 14 state-of-the-art VLMs indicates that current VLMs remain far from reliable for longitudinal medical reasoning, with an average overall accuracy of only \textbf{29.3\%} across task categories.


We also employ a stage-wise task decomposition that separates interval-level evidence articulation from final decision making, enabling a principled examination of how different task structures stress distinct aspects of model reasoning.
Through this analysis, we show that while many models are capable of articulating local interval-level changes when appropriately prompted, they frequently fail to enforce exclusivity, bind events into ordered temporal structures, or compose interval-level observations into coherent global trajectories. These findings highlight a fundamental limitation of current VLMs: the bottleneck lies not only in visual perception, but also in structured temporal decision-making over extended sequences.
In summary, our contributions are threefold:
\begin{itemize}[leftmargin=1.3em, labelsep=0.5em, topsep=0.25ex, itemsep=0.25ex, parsep=0pt]
    \item We introduce \textbf{MI-CXR}, a benchmark that systematically evaluates \emph{Temporal Event Localization}, \emph{Interval-wise Change Reasoning}, and \emph{Global Trajectory Summarization} over multi-interval CXR sequences.
    \item We formalize longitudinal CXR interpretation as a global reasoning problem grounded in realistic clinical workflows, emphasizing temporally structured constraints (ordering, exclusivity, and trajectory-level consistency).
    \item We present a stage-wise diagnostic framework that characterizes local and global reasoning failures in current VLMs, revealing a systematic gap where locally correct observations do not reliably yield correct longitudinal decisions.
\end{itemize}


\section{Related works}
\subsection{Medical Visual Question Answering for Chest X-ray}


Medical VQA for chest X-ray images has been widely studied as a benchmark for multimodal understanding in clinical imaging~\cite{slake, pmc-vqa, pubmed, EHRVQA, MIMIC_Ext_CXR_VQA, chexpert_plus}.
Early datasets such as VQA-RAD~\cite{vqa-rad} and PathVQA~\cite{pathvqa} focus on single-image settings, evaluating snapshot-level recognition of abnormalities, anatomical structures, and image attributes.
Subsequent work extends this paradigm to pairwise comparison settings, with benchmarks such as MIMIC-Diff-VQA~\cite{mimic-diff-vqa}, MMXU~\cite{mmxu}, and TemMed-Bench~\cite{temmed} targeting local changes between two visits.

Beyond two-image settings, a few recent benchmarks have begun to incorporate multi-visit CXR data, though with substantially different objectives from longitudinal reasoning evaluation. 
For example, LUNGUAGE~\cite{lunguage} focuses on report generation over image sequences, while CXReasonBench~\cite{cxreasonbench} introduces limited multi-timepoint question answering.
However, these benchmarks are not designed to explicitly evaluate long-horizon longitudinal reasoning over temporally ordered visual evidence.

Taken together, prior medical VQA benchmarks for chest X-rays fall short in evaluating long-horizon longitudinal reasoning: single-image and pairwise datasets are limited to short-term reasoning, while existing multi-visit benchmarks emphasize report generation or presence–absence judgments without probing temporal event ordering, recurrence, or resolution.
Moreover, they do not disentangle different stages of temporal reasoning, making it difficult to analyze where longitudinal inference breaks down.

\subsection{Longitudinal Modeling and Multi-visit Reasoning in Chest X-ray}
Recent studies have explored longitudinal modeling for CXR analysis by incorporating prior images, historical reports, and clinical context to improve diagnostic fidelity and report quality~\cite{zhang2025libraleveragingtemporalimages, cho2024pretrainingvisionlanguagemodeldifference, EKAID, qiu2021describinglocalizingmultiplechanges, zhang2024rexrankpublicleaderboardaipowered}.
Many of these approaches focus on radiology report generation, such as PriorRG~\cite{priorrg}, HERGen~\cite{hergen}, and MAIRA-2~\cite{maria2}, or on representation learning from longitudinal data, such as MLRG~\cite{mlrg}.

While these methods demonstrate the value of multi-visit information for modeling and generation, they are not designed to directly evaluate whether models can reason over longitudinal visual evidence.
In contrast, we focus on task-driven evaluation of longitudinal reasoning, highlighting a gap between existing longitudinal modeling approaches and the need for principled evaluation benchmarks.

\section{MI-CXR}
\label{sec:benchmark_construction}

\begin{figure*}[t!]
\centering
\includegraphics[width=\linewidth]{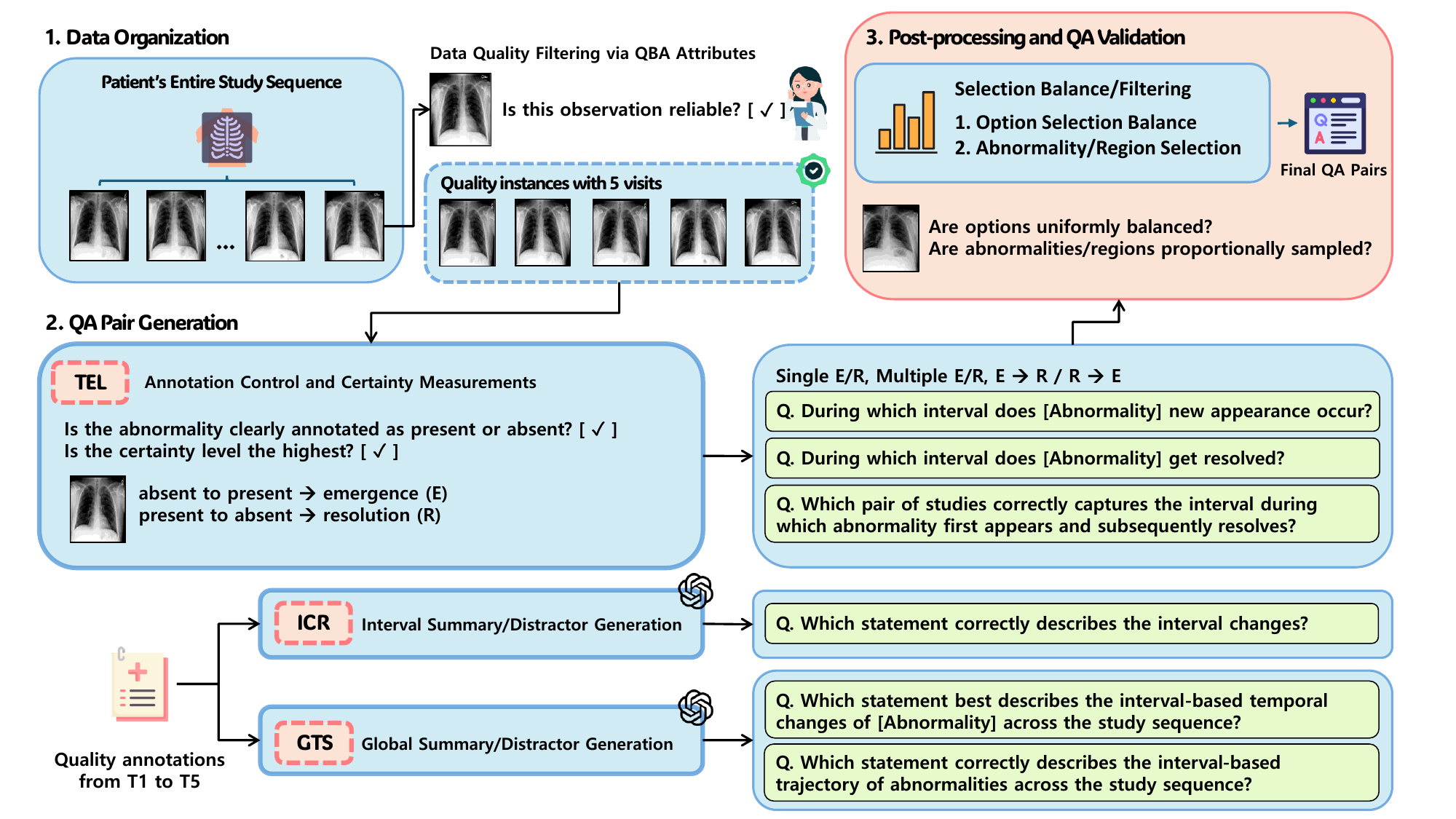} 
\caption{\textbf{Overview of MI-CXR construction.} We repurpose structured metadata from MIMIC-Ext-CXR-QBA and chest X-ray images from MIMIC-CXR-JPG to construct patient-level longitudinal timelines with at least five visits. After fixing the longitudinal cohort, multiple question types are instantiated from the same timelines to evaluate complementary longitudinal reasoning capabilities, including temporal event localization, interval-wise change reasoning, and global trajectory summarization.}
\label{fig:pipeline_fig_2}
\vspace{-0.5em}
\end{figure*}

We introduce \textbf{MI-CXR}, a new benchmark to evaluate global longitudinal reasoning over multi-visit CXR images.
MI-CXR comprises 5,311 multi-choice instances across three task families:
Temporal Event Localization (TEL), Interval-wise Change Reasoning (ICR), and Global Trajectory Summarization (GTS).
Each instance is built on a patient timeline consisting of five temporally ordered visits (i.e., five CXR studies), which yields four consecutive intervals for temporal reasoning, and consists of a CXR sequence, a natural-language question, and a set of answer options. 
\subsection{Task Description}

\paragraph{Temporal Event Localization (TEL)}
TEL requires identifying when clinically meaningful events, such as abnormality emergence or resolution, occur along a multi-visit timeline.
We consider three temporal patterns with increasing structural complexity: Single (E/R), Multiple (E/R), and ordered event patterns (E$\rightarrow$R or R$\rightarrow$E), which require reasoning under temporal ordering and exclusivity constraints.

\paragraph{Interval-wise Change Reasoning (ICR)}
ICR focuses on interpreting changes between consecutive visits.
Unlike standard pairwise comparison settings, the relevant interval is not specified in the question, requiring models to first localize the described change within the timeline before interpreting its semantic content.

\paragraph{Global Trajectory Summarization (GTS)}
GTS requires integrating evidence across all visits to characterize the overall disease course.
We instantiate both single- and multi-abnormality cases, where models must summarize or compare temporal trajectories grounded in interval-level observations.

\subsection{Benchmark Construction}
As illustrated in Figure~\ref{fig:pipeline_fig_2}, MI-CXR is constructed through a multi-stage pipeline including longitudinal cohort formation, annotation quality filtering, task instantiation, and post-processing. We follow the official MIMIC-CXR patient-level split, ensuring no patient appears across subsets.
\paragraph{Data Organization}
To achieve the high-quality and fine-grained assessment, we follow the multi-stage
construction pipeline by repurposing patient-study-timestamp metadata from MIMIC-Ext-CXR-QBA~\citep{mimicqba} and receiving the corresponding CXR images from MIMIC-CXR-JPG~\citep{mimiccxrjpg}.
We first organize studies into patient-wise longitudinal sequences by sorting study timestamps, ensuring that all images within a timeline correspond to the same patient and that visit ordering is consistent with the acquisition time.
We then fix the longitudinal cohort by retaining patients with at least five temporally ordered visits; for each eligible patient we construct a five-visit timeline. The inter-study intervals within five-visit windows span a wide range—from same-day follow-ups to multi-year longitudinal trajectories—reflecting realistic clinical monitoring patterns rather than artificially constrained scenarios. This design choice captures diverse longitudinal patterns encountered in practice (see Appendices~\ref{sec:source_metadata}--\ref{sec:excluded_cases} for the metadata mapping and cohort selection details).


Before the task instantiation, we apply a data quality filtering stage based on the quality attributes provided in MIMIC-Ext-CXR-QBA. Specifically, we retain only annotations that meet predefined quality thresholds across the annotated multiple dimensions, ensuring that all downstream summaries and questions are constructed from high-confidence radiologist-derived observations.

\paragraph{QA Pair Generation}
After fixing the longitudinal cohort and validating annotation quality, we instantiate three task families (TEL, ICR, GTS) from the same timelines to probe complementary aspects of longitudinal reasoning.
For each question, we generate a five-way option set consisting of one correct answer and multiple distractors.
Correct answers are constructed by recombining annotated findings into temporally coherent statements.
Distractors introduce controlled factual inconsistencies, such as incorrect temporal placement or change direction, while remaining annotation-grounded and clinically plausible (see Appendix~\ref{sec:question_templates} for question templates across each task).

Importantly, distractors are designed to remain annotation-grounded and avoid unsupported clinical inference, ensuring that incorrect options are plausible but definitively wrong under careful temporal reasoning.
Detailed generation procedure is depicted in Appendix~\ref{sec:llm_generation}.

\begin{table*}[t!]
\centering
\resizebox{\textwidth}{!}{
\begin{tabular}{llccccccc}
\toprule
\multirow{2}{*}{\textbf{Category}}
& \multirow{2}{*}{\textbf{Model}} 
& \multicolumn{3}{c}{\textbf{TEL}} 
& \textbf{ICR} 
& \multicolumn{2}{c}{\textbf{GTS}} 
& \multirow{2}{*}{\textbf{Overall}} \\
\cmidrule(lr){3-5} \cmidrule(lr){6-6} \cmidrule(lr){7-8}
& 
& \textbf{Single (E/R)}
& \textbf{Multiple (E/R)}
& \textbf{E$\rightarrow$R / R$\rightarrow$E }
& \textbf{--}
& \textbf{Single Abnormality}
& \textbf{Multi Abnormality}
&  \\

\midrule
\multirow{3}{*}{\textbf{Closed}}
& Claude Sonnet 4.5   & 0.226 & 0.222 & 0.243 & 0.442 & 0.389 & 0.292 & 0.315 \\
& Gemini 3.0 Pro      & 0.246 & 0.325 & 0.290 & 0.457 & 0.556 & 0.407 & 0.387 \\
& GPT-5.2             & \textbf{0.334} & \textbf{0.371} & \textbf{0.358} & 0.438 & \textbf{0.558} & 0.390 & 0.411 \\

\midrule
\multirow{7}{*}{\textbf{General}}
& InternVL3.5-8B      & 0.239 & 0.295 & 0.193 & 0.552 & 0.389 & 0.371 & 0.358 \\
& InternVL3.5-14B     & 0.248 & 0.266 & 0.223 & 0.293 & 0.374 & 0.276 & 0.281 \\
& InternVL3.5-38B     & 0.298 & 0.306 & 0.224 & \textbf{0.571} & 0.510 & \textbf{0.515} & \textbf{0.418} \\
& QwenVL3-8B          & 0.237 & 0.295 & 0.185 & 0.164 & 0.328 & 0.236 & 0.234 \\
& QwenVL3-32B         & 0.258 & 0.246 & 0.240 & 0.224 & 0.363 & 0.325 & 0.272 \\
& DeepSeek-VL-16B     & 0.223 & 0.124 & 0.200 & 0.186 & 0.160 & 0.187 & 0.181 \\
& IDEFICS2-8B         & 0.165 & 0.308 & 0.291 & 0.246 & 0.281 & 0.178 & 0.245 \\

\midrule
\multirow{4}{*}{\textbf{Medical}}
& Lingshu-7B          & 0.230 & 0.260 & 0.165 & 0.189 & 0.324 & 0.194 & 0.223 \\
& Lingshu-32B         & 0.221 & 0.247 & 0.214 & 0.167 & 0.388 & 0.290 & 0.247 \\
& MedGemma-4B         & 0.174 & 0.196 & 0.301 & 0.281 & 0.259 & 0.183 & 0.237 \\
& MedGemma-27B        & 0.215 & 0.351 & 0.254 & 0.429 & 0.255 & 0.214 & 0.299 \\

\bottomrule
\end{tabular}
}
\caption{\textbf{Baseline performance of the state-of-the-art VLMs on MI-CXR.} Results are reported across task families and question subtypes under single-step prompting. Overall low accuracy across models highlights the difficulty of long-horizon temporal diagnostic reasoning and motivates further analysis of underlying failure modes. See Appendix~\ref{sec:appendix_temperature} for results with different temperature setting for evaluated models.}
\label{tab:main_results}
\vspace{-0.5em}
\end{table*}

\paragraph{Post-processing and QA Validation}
After QA pair generation, we apply post-processing and validation to ensure balanced and reliable evaluation. All multiple-choice questions use a fixed option set (A–E). Correct answer positions are uniformly distributed across the options, preventing selection bias~\cite{positional, robust}. Additionally, abnormality types are sampled to match their overall frequency distribution, ensuring proportional representation across entities.

Both correct answers and distractors are validated using three annotation-aligned criteria: annotation coverage, change direction consistency, and context bound insurance. Correct answers must satisfy all criteria, while distractors must violate at least one factual criterion without introducing over-interpretation. The detailed validation protocol and result are provided in Appendix~\ref{sec:dataset_eval}.

Finally, starting from an initial pool of 11,234 candidate QA pairs constructed from the longitudinal cohort (patients with at least 5 CXR visits), we retain 5,311 high-quality longitudinal CXR QA instances after quality filtering and validation.
Detailed dataset statistics are presented in Appendix~\ref{sec:stat_break}.
\section{Experiments}
\subsection{Experimental Setup}
\label{sec:experimental_setup}

We evaluate 14 state-of-the-art VLMs on MI-CXR.
We adopt zero-shot prompting as the primary protocol to ensure reproducible, cross-model comparisons, since few-shot performance can be sensitive to exemplar selection and ordering, and constructing demonstrations for longitudinal medical reasoning risks unintended guidance.

Following our dataset design, the evaluation focuses on annotation-grounded temporal reasoning and does not provide free-text radiology reports or additional clinical context.
Unless otherwise stated, decoding is deterministic (temperature $=0$ and default settings for other sampling parameters under each provider’s recommended protocol).
Models are instructed to output exactly one choice among A--E.
We apply a single deterministic rule-based extraction procedure shared across all models to map outputs to a valid option; outputs that do not yield a valid option are counted as incorrect.
As the primary evaluation metric, accuracy is computed overall and per task family/subtype.

Evaluated models are grouped into three categories: closed-source general-purpose VLMs~\cite{openai2025gpt52, anthropic2024sonnet45, google2024gemini3}, open-source general-purpose VLMs~\cite{wang2025internvl3_5, qwen3technicalreport, wu2024deepseekvl2mixtureofexpertsvisionlanguagemodels, laurençon2024matters}, and medical-specialized VLMs~\cite{lasateam2025lingshugeneralistfoundationmodel, sellergren2025medgemmatechnicalreport}.
This grouping lets us examine whether domain specialization (or scale) is associated with longitudinal reasoning performance on MI-CXR.

\subsection{Baseline Performance}
\label{sec:diag_finding}


Table \ref{tab:main_results} summarizes baseline performance across all task families and question subtypes under single-step prompting. Because MI-CXR is a five-way multiple-choice benchmark, random guessing yields 20\% accuracy; we therefore emphasize both absolute accuracy and the margin over chance. Across model categories, even for large-scale or medically specialized models, we observe \textbf{consistently low performance and non-trivial variance across task families}, indicating that long-horizon temporal reasoning over multi-visit CXR remains challenging for current VLMs rather than being driven by a single pathological task design.


\subsubsection{Task-wise Comparison}
\paragraph{Temporal Event Localization (TEL)}
Overall performance on TEL is uniformly low across all TEL subtypes.
Accuracy remains modest for Single E/R questions, indicating that even basic temporal grounding over extended timelines is unreliable.
Similarly low performance is observed for Multiple E/R and E→R / R→E questions, suggesting that TEL remains challenging across all task formulations, regardless of subtype-specific complexity.

\paragraph{Interval-wise Change Reasoning (ICR)}
Overall performance on ICR remains consistently low across models, indicating that the task difficulty extends beyond visual change recognition to the challenge of selecting the correct interval among multiple plausible candidates within a longer timeline.
Nevertheless, several model families exhibit comparatively stronger performance, including the InternVL3.5 family (with the exception of the 14B variant), all closed-source models, and MedGemma, which consistently outperform the remaining open-source baselines on this task.

\paragraph{Global Trajectory Summarization (GTS)}
Performance on GTS differs by question structure.
Across most evaluated models, Single Abnormality questions tend to yield higher accuracy than Multi Abnormality questions, while a small number of models deviate from this pattern.
This indicates that introducing multiple abnormalities generally increases reasoning difficulty by expanding the space of competing global trajectories.

\subsubsection{Model Category Comparison}

Across task families, closed-source models generally achieve higher average performance, particularly on ICR and GTS, but this trend is not uniform across all closed models or task types, and none demonstrate consistently robust longitudinal reasoning across TEL, ICR, and GTS.

Open-source models exhibit substantial intra-family variance: while larger variants (e.g., InternVL3.5-38B) often outperform their smaller counterparts, this scaling effect is inconsistent, and several models still struggle on complex TEL subtypes and multi-abnormality GTS settings.

Medical-specialized VLMs show mixed behavior, occasionally matching or exceeding general-purpose models on specific tasks (e.g., GTS Single Abnormality or ICR for MedGemma-27B), but do not exhibit systematic advantages across task families or within their own model families.

Overall, these results indicate that neither model category nor parameter scale alone reliably predicts longitudinal reasoning performance, and that temporal reasoning failures persist across architectures and domains.



\subsection{Capability-Aligned Task Decomposition}
\label{sec:capability_decomposition}

The baseline results indicate that directly prompting models to reason over five images in a single step yields limited performance. However, aggregate accuracy does not clarify whether failures stem from missing local visual evidence or from difficulties in integrating such evidence into a globally consistent decision. To probe this distinction, we analyze model capabilities under a controlled capability-aligned task decomposition.

\subsubsection{Probing Local Interval Reasoning Capability}
We first examine whether models can correctly reason about visual changes when the relevant temporal interval is explicitly specified.
To this end, we construct an ICR variant dataset in which each question focuses on a predefined pair of visits.
This formulation isolates local interval reasoning by removing the need for interval selection.

\begin{table}[h!]
\centering
\small
\begin{tabular}{llc}
\toprule
\textbf{Category} & \textbf{Model} & \textbf{ICR Variant} \\
\midrule
\multirow{3}{*}{\textbf{Closed}}
& Claude Sonnet 4.5      & 0.601 \\
& Gemini 3.0 Pro         & 0.743 \\
& GPT-5.2                & \textbf{0.765} \\
\midrule
\multirow{7}{*}{\textbf{General}}
& InternVL3.5-8B         & 0.667 \\
& InternVL3.5-14B        & 0.661 \\
& InternVL3.5-38B        & 0.634 \\
& QwenVL3-8B              & 0.590 \\
& QwenVL3-32B             & 0.601 \\
& DeepSeek-VL-16B        & 0.284 \\
& IDEFICS2-8B             & 0.448 \\
\midrule
\multirow{4}{*}{\textbf{Medical}}
& Lingshu-7B              & 0.585 \\
& Lingshu-32B             & 0.705 \\
& MedGemma-4B             & 0.617 \\
& MedGemma-27B            & 0.705 \\
\bottomrule
\end{tabular}
\caption{\textbf{Performance comparison on ICR Variant across model categories.}}
\label{tab:icr_easy_category}
\end{table}

Table~\ref{tab:icr_easy_category} reports model performance on the ICR variant.
Results show that models achieve substantially higher accuracy on this ICR variant compared to ICR in the main benchmark, suggesting that many VLMs are capable of interpreting interval-level changes when the temporal scope is constrained. Detailed dataset construction and performance statistics for this variant are provided in Appendix~\ref{sec:icr_variant}. 

\subsubsection{Stage-wise Diagnostic Probing}
\label{sec:stagewise_probing}
Motivated by this observation, we further evaluate models using a stage-wise inference protocol aligned with their apparent capabilities. In the first stage, models are prompted to generate structured, interval-wise descriptions of visual changes between consecutive visits. In the second stage, models answer the original question based solely on these intermediate descriptions.

By separating local evidence articulation from decision making, we aim to assess whether models already possess useful interval-level understanding that is not effectively utilized under single-step prompting. See Appendix~\ref{sec:appendix_eval_protocol} for detailed probing construction.

\begin{figure}[h!]
\vspace{-1em}
\includegraphics[width=0.98\columnwidth]{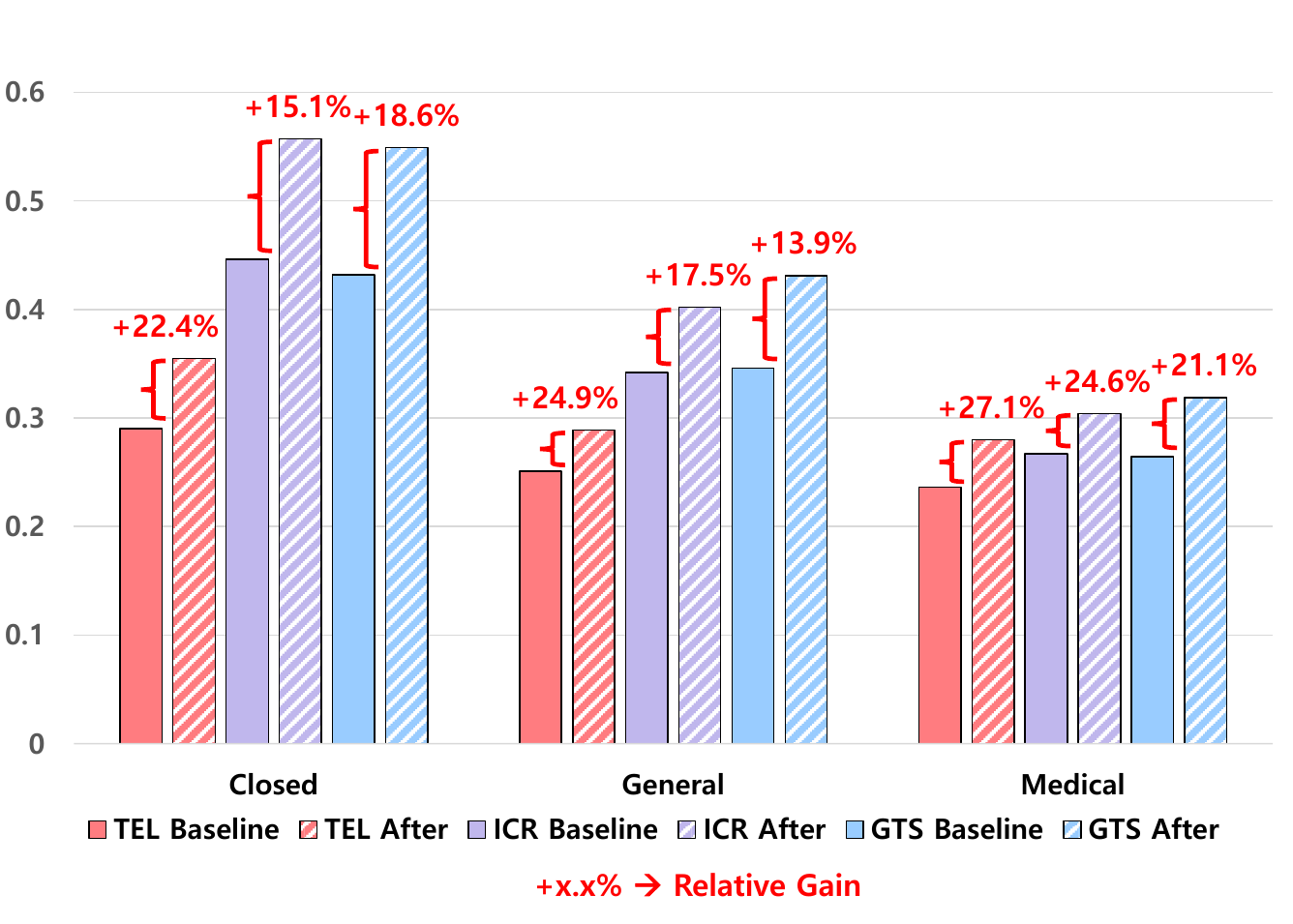} 
\caption{\textbf{Performance under capability-aligned task decomposition.}
Models are evaluated using a stage-wise inference protocol that separates interval-level evidence articulation from final decision making. 
}
\label{fig:pipeline_fig_3}
\end{figure}

As shown in Figure \ref{fig:pipeline_fig_3}, this capability-aligned decomposition consistently improves performance across models\footnote{We attempted to evaluate DeepSeek-VL under the same evaluation protocol; however, it did not consistently produce valid stage-wise intermediate outputs and was therefore omitted from the reported results.} and task types.
The gains indicate that prompting models to explicitly reason about each interval reduces interference among competing temporal hypotheses and allows them to better leverage their local comparison capabilities.

\section{Error Patterns in Longitudinal Grounding}
\label{sec:error_analysis}

To better understand the limitations of current VLMs on longitudinal medical reasoning, we analyze error patterns across task categories.
A central finding is that most failures do not stem from misperceiving individual images or short-term changes, but from breakdowns in temporal decision-making when models must reason over multiple dependent visits.
Locally plausible interpretations often fail to propagate into coherent global conclusions, revealing systematic weaknesses in how temporal evidence is selected, prioritized, and integrated.

Rather than treating errors as isolated mistakes, we analyze how different task families expose distinct but related failure modes.
By jointly examining the intermediate interval-level outputs from the stage-wise inference protocol (Section~\ref{sec:stagewise_probing}) and final model decisions, we identify where errors arise in the reasoning pipeline and why they manifest differently across tasks.

\subsection{Failures in Temporal Event Localization}

A common failure mode in TEL is that models recognize the presence of an event but struggle to localize it precisely. 
Under the stage-wise inference protocol, models often correctly describe the emergence or resolution of an event across multiple adjacent intervals, but fail to identify a single interval as temporally decisive, as illustrated in Figures~\ref{fig:f_tel_single}--\ref{fig:f_tel_erre} from Appendix~\ref{sec:failure_tel_abstention}.
Importantly, stage-wise analysis shows that models frequently treat multiple intervals as equally plausible candidates for the target event.
When tasks require selecting the earliest occurrence or a unique decisive interval, models often fail to enforce these constraints, leading to incorrect predictions.

These patterns indicate that TEL is challenging because it requires resolving competing temporal hypotheses.
Current models lack robust mechanisms for prioritizing one interval over others when multiple points along the timeline appear consistent with the event, resulting in systematic breakdowns in temporal grounding.

\subsection{Failures under Interval Ambiguity in Change Reasoning}



Interval-wise Change Reasoning (ICR) focuses on interpreting visual changes between consecutive visits and committing to a specific interval-level interpretation.
Errors in this task family frequently arise when visual differences are subtle or when the visual evidence supporting a particular change is weak.
In such cases, ambiguity is not due to the absence of observable findings, but rather to marginal differences that do not clearly support a single directional interpretation.

As illustrated in Figure~\ref{fig:f_icr} from Appendix~\ref{sec:failure_icr_forced},  stage-wise intermediate outputs often describe interval-level changes using hedged or non-committal language (e.g., ``only marginal and ambiguous change''), indicating that multiple interpretations remain plausible.
Despite this ambiguity, models tend to
overcommit to a specific abnormality and direction of change without sufficient supporting evidence.

Stage-wise analysis shows that when interval-level signals are marginal, models often either misinterpret the direction or magnitude of change, or collapse uncertainty into a single hypothesis during answer selection.
This results in forced-choice errors, where definitive decisions are made despite the absence of a clearly supported interval-level change.

These failures highlight the interaction between fine-grained visual discrimination and decision commitment.
When correctness depends on subtle distinctions across short intervals, the requirement to select a single answer amplifies uncertainty and exposes limitations in how models balance caution against decisiveness.

\subsection{Failures in Global Evidence Integration and Trajectory Reasoning}

Global Trajectory Summarization (GTS) presents the greatest challenge among the task families, as it requires integrating evidence distributed across all visits into a coherent interpretation of disease progression.
Errors in this setting are dominated by failures in composing interval-level information and maintaining consistency at the trajectory level.

Even when interval-level descriptions are largely correct, models often fail to reconcile these observations into a single globally consistent summary as shown in Figures~\ref{fig:f_gts_single}--\ref{fig:f_gts_multi} from Appendix~\ref{sec:gts_failure}.
Stage-wise analysis shows that local interpretations do not reliably propagate into final decisions, and mild inconsistencies or early misjudgments can cascade into incorrect global conclusions.

These patterns point to a fundamental limitation in temporal evidence integration.
As the temporal scope increases and correctness depends on joint reasoning across multiple intervals, models struggle to maintain coherence over the full sequence.
This fragility explains why performance degrades most sharply in tasks that require holistic trajectory reasoning rather than isolated or short-horizon comparisons.

Taken together, these error patterns suggest that the primary bottleneck in longitudinal medical reasoning lies not only in visual perception, but also in temporal decision-making under dependency.
Different task families expose distinct aspects of this limitation, including the ability to assign decisiveness to specific intervals, commit under ambiguity, and integrate distributed evidence into stable global interpretations.
By revealing where and why temporal reasoning breaks down, our analysis provides guidance for the development of future models capable of reliable longitudinal inference.
A quantitative breakdown of these error types across task families and model categories is provided in Appendix~\ref{sec:error_distribution}.
\section{Robustness Analysis}
\label{sec:robustness_analysis}

The error patterns identified in Section~\ref{sec:error_analysis} suggest that these failures may stem either from intrinsic limitations in longitudinal reasoning or from artifacts related to task-format unfamiliarity and prompt sensitivity. We investigate these possibilities through two complementary analyses.

\paragraph{Report Generation Format} To examine whether 
failures persist beyond the multiple-choice format, 
we conducted a pilot study in which representative 
models generated free-form interval-based summaries 
for GTS-type questions (Appendix~\ref{sec:report_generation}). All models 
achieved modest scores across natural language generation (NLG) metrics (Table~\ref{tab:report_generation_results}).
Qualitative failure modes closely aligned with 
those observed in the structured MCQ setting: 
models produced locally plausible interval 
descriptions but failed to compose them into 
coherent global trajectories. This convergence 
across evaluation formats indicates that the 
primary bottleneck lies in longitudinal evidence 
integration rather than in output generation 
mechanics, reinforcing the MCQ formulation as a 
reproducible and format-independent diagnostic 
tool.

\paragraph{Prompting Strategies} We evaluated two 
prompting variants beyond the zero-shot baseline: 
reasoning-style guidance \cite{NEURIPS2022_8bb0d291}, and question type-matched 1-shot demonstration. 
As shown in Appendix~\ref{sec:prompting_analysis}, neither of these variants 
yielded consistent performance improvements. 
Notably, 1-shot prompting produced slight 
degradation across all three representative models 
(Table~\ref{tab:fewshot_prompting}), suggesting that introducing 
demonstration examples does not alleviate 
cross-interval integration demands and may in fact 
dilute temporal attention by expanding the visual 
context. Across all prompting variations, the 
qualitative failure patterns identified in Section~\ref{sec:error_analysis} 
remain intact, confirming that they do not stem from task-format ambiguity or prompt sensitivity.
\section{Conclusion}
In this work, we introduce MI-CXR, a benchmark for longitudinal medical visual question answering that targets temporal reasoning over multi-visit CXR sequences.
By formulating longitudinal interpretation as Temporal Event Localization, Interval-wise Change Reasoning, and Global Trajectory Summarization, our benchmark moves beyond single-image and pairwise evaluations toward clinically grounded assessment of visual reasoning over time.
Our evaluation shows that current VLMs struggle consistently across all longitudinal task categories.
These failures are not primarily due to deficient visual perception, but rather to limitations in temporal decision-making.

We hope this benchmark will serve as a diagnostic tool for evaluating longitudinal reasoning in multimodal medical AI system and motivate future work on representations and inference mechanisms that better support structured temporal reasoning in medical imaging.

\newpage
\section*{Limitations}
Our analysis is centered on visual and decision-level reasoning, and does not incorporate complementary clinical context such as laboratory values or textual reports, which often inform longitudinal interpretation in practice. Also, the stage-wise evaluation framework enables diagnostic analysis of temporal reasoning failures, it does not reveal the internal mechanisms by which models process temporal information.
We view these limitations as opportunities for future work to extend longitudinal reasoning benchmarks to broader modalities, longer temporal horizons, and richer multimodal settings.

\section*{Ethical Consideration}

MI-CXR is a benchmark designed to evaluate longitudinal reasoning capabilities of vision--language models on chest X-ray sequences.
It is intended solely for research and evaluation purposes, and not for clinical deployment or decision-making.

\paragraph{Data Source and Privacy}
MI-CXR is constructed by repurposing publicly available datasets, MIMIC-CXR-JPG and MIMIC-Ext-CXR-QBA, which are distributed via PhysioNet under the PhysioNet Credentialed Health Data License (Version 1.5.0). All patient identifiers are removed, and no attempt is made to re-identify individuals.
MI-CXR does not introduce new annotations that could enable re-identification, nor does it modify the original data in a manner that weakens the privacy guarantees of the source datasets.
All results are reported in aggregate form, and no individual-level information is disclosed.

\paragraph{Clinical Safety and Misuse Risks}
Although MI-CXR involves medical images and clinically grounded questions, it does not provide diagnostic recommendations or treatment guidance.
The benchmark evaluates whether models can reason over temporally ordered visual evidence, not whether they can make correct clinical decisions.
We strongly discourage the use of models evaluated on MI-CXR for autonomous clinical interpretation or medical decision-making without rigorous validation, regulatory approval, and human oversight.
Incorrect longitudinal reasoning—such as misidentifying disease onset, resolution, or recurrence—could lead to harmful conclusions if misused in clinical settings.

\paragraph{Transparency and Reproducibility}
We aim to promote transparency and reproducibility by clearly documenting the benchmark construction process, task definitions, and evaluation protocols.
MI-CXR is released to support responsible research on longitudinal medical reasoning and error analysis.
We encourage future work to build upon this benchmark to develop more robust, interpretable, and clinically aligned longitudinal reasoning models, while adhering to ethical standards for medical AI research.

\section*{Acknowledgements}
This work was supported in part by National Research Foundation of Korea (NRF) grants (RS-2024-00414981), Institute of Information \& Communications Technology Planning \& Evaluation (IITP) grants (RS-2024-00397085, RS-2021- II211343), and the Health and Medical R\&D Program of the Ministry of Health and Welfare (RS2025-25455059). This research was also conducted as part of the Creative-Pioneering Researchers Program and the Bio-Connect Program through the Bio-MAX Institute at Seoul National University, and supported by the “Advanced GPU Utilization Support Program” funded by the Government of the Republic of Korea (Ministry of Science and ICT) (RQT-25-090156). J. Do is with ASRI, Seoul National University.

\bibliography{custom}
\clearpage











\appendix
\clearpage

\twocolumn[{
\begin{center}

\centering
\small
\setlength{\tabcolsep}{4pt}
\begin{tabular}{p{3cm} l p{7.4cm}}
\toprule
\textbf{Category} & \textbf{Field} & \textbf{Description} \\
\midrule
\multirow{4}{*}{Study-level Metadata}
& \texttt{patient\_id} & Unique patient identifier used for longitudinal aggregation. \\
& \texttt{study\_id} & Unique imaging study identifier. \\
& \texttt{study\_index} & Relative temporal index of the study within a patient timeline. \\
& \texttt{timestamp} \& \texttt{days\_since\_prev} & Temporal information used to verify chronological ordering. \\
\midrule
\multirow{3}{*}{Image Metadata}
& \texttt{image\_id} & Unique identifier for each chest radiograph. \\
& \texttt{view\_position} & Acquisition view (e.g., PA, lateral). \\
& \texttt{image\_size} & Image resolution and dimensions. \\
\midrule
\multirow{5}{*}{Observation Annotation}
& \texttt{obs\_entities} & Radiological abnormality entities (e.g., cardiomegaly, pleural effusion). \\
& \texttt{obs\_categories} & High-level category (e.g., disease, anatomical finding). \\
& \texttt{regions} \& \texttt{laterality} & Anatomical localization and laterality information. \\
& \texttt{changes} & Explicit temporal change labels (emergence, resolution, improvement, persistence). \\
& \texttt{change\_sentence} & Natural language description of the annotated change. \\
\midrule
\multirow{4}{*}{Quality Indicators}
& \texttt{certainty} & Certainty level of the observation annotation. \\
& \texttt{obs\_quality} & Observation-level quality scores for entity, region, and change extraction. \\
& \texttt{study\_quality} & Overall study-level annotation quality. \\
& \texttt{localization\_quality} & Quality of spatial grounding for annotated regions. \\
\bottomrule
\end{tabular}
\captionof{table}{\textbf{Metadata fields utilized from MIMIC-Ext-CXR-QBA for dataset construction.}}
\label{tab:metadata_fields}
\end{center}
\vspace{1em}
}]

\section{Details of MI-CXR}
\label{sec:construction_details}

\subsection{Source Datasets and Metadata Fields}
\label{sec:source_metadata}

MI-CXR is constructed by integrating chest radiographs from the MIMIC-CXR-JPG dataset~\citep{mimiccxrjpg} with structured, high-resolution annotations provided by MIMIC-Ext-CXR-QBA~\citep{mimicqba}.
We exclusively rely on the structured metadata and scene graph annotations released by MIMIC-Ext-CXR-QBA, and do not directly use free-text radiology reports during question construction.

Each imaging study is associated with a metadata file that contains patient-level and study-level information, including patient identifiers, study identifiers, relative temporal indices within a patient timeline, acquisition timestamps, and detailed image attributes such as view position (e.g., PA, lateral), patient orientation, and image resolution (Table~\ref{tab:metadata_fields}).
These metadata fields enable unambiguous patient--study mapping and temporal ordering across longitudinal imaging sequences.

In addition, each study is accompanied by a scene graph annotation that encodes radiological observations in a fully structured form.
Each observation specifies the abnormality entity (e.g., cardiomegaly, pleural effusion), anatomical regions and laterality, categorical labels (e.g., disease or anatomical finding), and explicit temporal change types such as emergence, resolution, improvement, or persistence.
Importantly, temporal changes are directly annotated in the scene graph rather than inferred post hoc from report text.

To ensure annotation reliability, the scene graph further provides multiple quality indicators at both the observation and study levels, including entity extraction quality, region localization quality, change annotation quality, and overall study-level quality scores.
We utilize these quality attributes to filter out uncertain or low-confidence annotations during dataset construction, retaining only observations with certain certainty labels and sufficient extraction quality.

Finally, we aggregate study-level scene graph annotations into patient-level temporal sequences, which serve as the foundation for subsequent sliding-window generation and question formulation.
This design allows our benchmark to focus on explicit, annotation-grounded temporal reasoning rather than implicit report interpretation.

\begin{table*}[t]
\centering
\small
\begin{tabular}{c c p{10cm}}
\toprule
\textbf{\# Visits ($T$)} & \textbf{\# Intervals ($T{-}1$)} & \textbf{Supported Reasoning Capabilities} \\
\midrule
2 & 1 & Single before--after comparison; no temporal pattern modeling. \\

3 & 2 & Single change detection; vulnerable to noise and transient fluctuations. \\

4 & 3 & Limited sequencing of changes; insufficient for global trajectory reasoning. \\

\textbf{5} & \textbf{4} & 
\textbf{Multiple emergence/resolution events; E→R or R→E transitions; interval-wise reasoning; global trajectory summarization}. \\

$>5$ & $>4$ & Longer trajectories with increased redundancy; handled via sliding-window generation. \\

\bottomrule
\end{tabular}
\caption{\textbf{Relationship between the number of patient visits and the temporal reasoning capabilities supported in the benchmark.}}
\label{tab:visit_threshold}
\end{table*}

\subsection{Patient--Study Mapping and Temporal Ordering}
\label{sec:patient_study_mapping}

\begin{figure}[h]
\centering
\begin{lstlisting}
patient_id: p10000980
study_sequence:
  - s50984512
  - s50984733
  - s50984901
  - s50985099
  - s50985321

studies:
  s50985099:
    - obs_entities: [pulmonary edema]
      changes: [resolution]
    - obs_entities: [cardiomegaly]
      changes: [improvement]
    - obs_entities: [pleural effusion]
      changes: [resolution]
\end{lstlisting}
\caption{\textbf{Example of a patient-level temporally ordered study sequence constructed from scene graph annotations.}
Each study represents a single clinical visit and aggregates all associated observations and temporal change labels.}
\label{fig:patient_yaml_example}
\end{figure}

All temporal reasoning tasks in our benchmark are constructed at the patient-level, where each patient is represented by an ordered sequence of imaging studies.
We aggregate studies using unique patient identifiers, and treat each study as a single clinical visit, regardless of the number of associated images (e.g., postero-anterior and lateral views).

Temporal ordering within each patient timeline is determined primarily by the \texttt{study\_index} field provided in the metadata, which encodes the relative chronological position of each study for a given patient.
This ordering is further validated using timestamp-related fields, including acquisition time and elapsed time since the previous study, to ensure temporal consistency.
Studies with ambiguous or inconsistent temporal information are excluded prior to timeline construction.

Each study thus corresponds to a discrete time point in the patient’s longitudinal trajectory, and may include multiple radiographs acquired during the same visit.
All abnormality observations and temporal change annotations are associated with the study-level time point, rather than individual images, to avoid artificial fragmentation of clinical events.

Following temporal ordering, each patient is represented as a strictly ordered sequence of studies:
\[
\texttt{Patient } p \rightarrow [s_1, s_2, \dots, s_T],
\]
where $T$ denotes the total number of valid visits for the patient.
These ordered patient-level study sequences serve as the fundamental input for subsequent filtering by minimum visit length, sliding window generation, and question construction.

By explicitly enforcing patient-level aggregation and unambiguous temporal ordering prior to question generation, our benchmark ensures that all temporal reasoning tasks are grounded in well-defined and clinically coherent longitudinal trajectories.

\subsection{Minimum Visit Threshold}
\label{sec:min_visit}

We restrict our dataset to patients with at least five valid imaging studies.
This minimum visit threshold is not chosen heuristically, but is a structural requirement imposed by the temporal reasoning tasks targeted in our benchmark (Table~\ref{tab:visit_threshold}).

A patient with $T$ longitudinal visits yields $T-1$ consecutive temporal intervals.
When $T < 5$, the resulting temporal context is insufficient to support well-defined temporal reasoning beyond trivial before--after comparisons.
In particular, timelines with two or three visits only permit isolated change detection and cannot disambiguate transient fluctuations from sustained disease progression or resolution.

With fewer than four intervals, it is not possible to reliably define temporal patterns involving multiple change events, such as repeated emergence or resolution, transitions between emergence and resolution (e.g., E→R or R→E), or persistent abnormalities followed by delayed resolution.
These patterns form the core of our question types.

By enforcing a minimum of five visits, each patient timeline contains four consecutive intervals, which is the smallest temporal span that enables:
\begin{itemize}[leftmargin=1.3em, labelsep=0.5em, topsep=0.25ex, itemsep=0.25ex, parsep=0pt]
    \item interval-wise reasoning over multiple adjacent changes,
    \item differentiation between temporary improvement and true resolution, and
    \item global summarization of abnormality trajectories across the entire observation window.
\end{itemize}
This temporal depth is essential for avoiding ill-posed questions that admit multiple valid interpretations.

Although this constraint reduces the total number of eligible patients, it substantially improves the semantic validity and clinical coherence of the resulting benchmark instances.
By excluding short or incomplete timelines, we ensure that each question is grounded in a longitudinal trajectory with sufficient temporal context to support unambiguous reasoning.

We note an additional practical consideration related to long temporal inputs.
As the number of visits increases, both open-source and closed-source vision--language models are required to process longer image sequences and more complex contextual information.
In practice, we observe that excessively long visit sequences may lead to increased inference instability, such as truncated responses or malformed outputs, even for large closed-source models.

Importantly, this observation does not motivate our minimum visit threshold, which is determined solely by the structural requirements of the targeted temporal reasoning tasks.
Rather, limiting the visit length helps avoid pathological failure cases during large-scale evaluation and ensures consistent benchmark execution across diverse model families.

\subsection{Sliding Window Generation}
\label{sec:sliding_window}








\begin{table}[H]
\centering
\small
\label{tab:sliding_window}
\begin{tabularx}{\columnwidth}{@{}l X@{}}
\toprule
\textbf{Design Aspect} & \textbf{Specification} \\
\midrule
Window size & 5 consecutive studies (visits) \\
Stride & 1 (overlapping windows) \\
Temporal constraint & Study indices must be strictly contiguous \\
Time granularity & Study-level (visit-level), not image-level \\
Reuse across tasks & Identical windows used for TEL, ICR, and GTS \\
Purpose & Normalize temporal context and enable consistent evaluation \\
\bottomrule
\end{tabularx}
\caption{\textbf{Sliding window generation strategy for patient timelines.}}
\vspace{-1em}
\end{table}

Patients with more than five valid imaging studies are decomposed into multiple fixed-length temporal windows using a sliding window strategy.
Each window consists of five consecutive studies, corresponding to five temporally ordered clinical visits, and serves as the basic unit for question construction.
Windows are generated with a stride of one, such that a patient with $T$ visits yields up to $T-4$ overlapping windows. To preserve temporal coherence, we only retain windows in which the study indices form a strictly contiguous sequence. This constraint ensures that no visits are skipped within a window and that all temporal intervals represent consecutive clinical observations.

All windows are constructed at the study (visit) level rather than the image level. Each study within a window may include multiple radiographs acquired during the same visit, which are jointly associated with the corresponding time point. Temporal change annotations are therefore aligned to visit-level intervals (e.g., $T_1 \rightarrow T_2$), avoiding artificial fragmentation of clinical events. Importantly, sliding windows are generated independently of the downstream question types. The same set of windows is reused across all task categories, including Temporal Event Localization, Interval-wise Change Reasoning, and Global Trajectory Summarization. Task-specific questions are subsequently instantiated by analyzing the temporal change patterns observed within each window.

This sliding window formulation allows the benchmark to leverage long patient histories while maintaining a fixed and controlled temporal context for each problem instance.
It also prevents data imbalance arising from variable sequence lengths and enables consistent evaluation across diverse model families.

\subsection{Excluded Cases}
\label{sec:excluded_cases}

\subsubsection{Window-level Filtering}
\label{sec:window_filtering}

Prior to question construction, we apply a series of filters at the window level to ensure temporal coherence and annotation completeness.
First, sliding windows are required to consist of strictly contiguous study indices, such that no visits are skipped within a window.
This constraint guarantees that all temporal intervals correspond to consecutive clinical observations.

Second, only observations annotated with \texttt{certain} certainty and belonging to valid disease-related categories are retained.
Windows in which an abnormality is absent or ambiguously annotated at any visit are excluded from further consideration.
As a result, each retained window--abnormality pair represents a complete and well-defined temporal sequence without missing states.

These window-level filters remove ill-posed or temporally ambiguous inputs before any task-specific questions are instantiated.

\subsubsection{Question-level Filtering and Balancing}
\label{sec:question_filtering}

After generating all candidate questions from valid windows, we apply additional filtering and sampling procedures at the question level to control dataset balance and difficulty.
This stage does not remove ill-defined questions, but rather enforces distributional constraints to prevent biased or degenerate evaluation.

Specifically, we impose per-question-type quotas to ensure balanced coverage across different temporal reasoning categories.
We further limit the proportion of questions associated with any single abnormality, preventing over-representation of common findings.
In addition, we cap the fraction of questions whose correct answer corresponds to a null option (e.g., ``none of the above''), which is known to induce shortcut strategies.

Finally, we regulate the distribution of window-level certainty patterns, such as windows containing uniformly positive or uniformly negative abnormality states.
These question-level constraints collectively improve benchmark robustness while preserving the semantic validity of each individual question.

\subsection{LLM-assisted Question Text Generation}
\label{sec:llm_generation}

\begin{table}[h]
\vspace{-0.5em}
\centering
\small
\label{tab:llm_role}
\begin{tabular}{ll}
\toprule
\textbf{Component} & \textbf{LLM Involvement} \\
\midrule
Temporal ordering & None (metadata-driven) \\
Abnormality detection & None (metadata-driven) \\
Change type labeling & None (annotated) \\
Interval/global summaries & Language realization only \\
Distractor generation & Rule-based flips only \\
Answer correctness & Deterministic, rule-based \\
\bottomrule
\end{tabular}
\caption{\textbf{LLM Involvement in question generation.}}
\vspace{-0.5em}
\end{table}

GPT-5.1~\citep{openai2025gpt51} is used in our benchmark solely to generate natural language question texts and answer options from pre-defined structured annotations.
All temporal ordering, abnormality identification, and change labels are determined prior to LLM invocation using scene graph annotations and rule-based logic.

For interval-wise and global trajectory summarization tasks, the LLM receives as input a structured representation of abnormality states and annotated temporal changes.
Its role is limited to verbalizing this information into concise natural language summaries under strict constraints.
Specifically, the model is instructed to preserve entity names, temporal intervals, and laterality, while refraining from introducing clinical interpretation, diagnostic inference, or additional findings.

To construct multiple-choice questions, incorrect answer options are generated by applying controlled semantic flips to change-type descriptors (e.g., ``improves'' versus ``worsens'', ``resolves'' versus ``persists'').
The LLM is explicitly constrained to modify only the taxonomy phrase, while keeping the temporal structure, entity references, and sentence format unchanged.
This design ensures that all distractors remain medically plausible yet objectively incorrect with respect to the underlying annotations.

Importantly, the LLM is never used to determine the correctness of answers, infer temporal relationships, or resolve ambiguities in the source data.
All correctness labels are derived deterministically from the original annotations.
By restricting the LLM to surface-level language generation, we avoid confounding benchmark difficulty with implicit model reasoning during dataset construction.

\subsubsection{Prompt Templates for LLM-assisted Question Generation}
\label{sec:prompt_templates}



We disclose all prompt templates used for LLM-assisted question text generation to ensure full reproducibility (Figures~\ref{fig:prompt_correct_interval}--\ref{fig:prompt_incorrect_global}).
Specifically, single-entity and multi-entity interval summary questions are generated using the interval-level prompt templates, while single-entity and multi-entity global summary questions are generated using the global-level prompt templates, each in both correct and incorrect variants.
All prompts are used exclusively for surface-level language realization from pre-defined structured annotations, and the LLM is never permitted to determine temporal order, abnormality presence, change types, or answer correctness.

All dataset construction steps are deterministic. Sliding window generation, filtering criteria, and correctness labels are entirely rule-based. LLM-assisted text generation is executed with fixed decoding settings and used only for surface-level realization. The final dataset is released as a static benchmark and does not require LLM access for evaluation.

\begin{figure}[h!]
\centering
\small
\begin{Verbatim}[fontsize=\scriptsize, frame=single]
You are a medical language assistant.

Your task is to generate a concise interval-based temporal
summary for a radiologic abnormality across an image sequence.

Rules:
- Describe the abnormality at the entity level.
- Include laterality ONLY if explicitly present.
- Mention all relevant intervals in temporal order.
- Use no more than ONE sentence.
- Use at most ONE semicolon.
- Do NOT infer etiology or diagnosis.
- Do NOT add information not present in the input.
\end{Verbatim}
\vspace{-1em}
\caption{\textbf{Correct interval summary prompt.}}
\label{fig:prompt_correct_interval}
\end{figure}

\begin{figure}[h!]
\centering
\small
\begin{Verbatim}[fontsize=\scriptsize, frame=single]
You generate incorrect but medically plausible interval-based
summaries for a single abnormality.

Rules:
- Use ONLY semantic change-type flips.
- Keep interval positions unchanged.
- Keep laterality unchanged if present.
- Do NOT match the correct summary.
- One sentence, at most one semicolon.
\end{Verbatim}
\vspace{-1em}
\caption{\textbf{Incorrect interval summary prompt.}}
\end{figure}

\begin{figure}[h!]
\centering
\small
\begin{Verbatim}[fontsize=\scriptsize, frame=single]
You are a medical language assistant.

You generate correct interval-based temporal summaries for
multiple radiologic abnormalities independently.

CRITICAL REQUIREMENTS:
- You MUST generate exactly ONE summary for EVERY abnormality
  provided.
- DO NOT omit any abnormality under any circumstance.
- Even if an abnormality shows no change, remains stable, or
  is normal, you MUST explicitly state that it remains
  stable or unchanged.
- The output JSON MUST contain exactly the same set of
  abnormality keys as the input abnormalities.

CONTENT RULES:
- Each summary must describe ONLY the specified abnormality.
- Mention all relevant intervals in temporal order.
- Use ONE sentence only.
- Use at most ONE semicolon.
- Include laterality ONLY if explicitly stated.
- Do NOT infer diagnosis or clinical implication.
\end{Verbatim}
\vspace{-1em}
\caption{\textbf{Correct global summary prompt (Multi-Entity).}}
\end{figure}

\begin{figure}[h!]
\centering
\small
\begin{Verbatim}[fontsize=\scriptsize, frame=single]
You generate incorrect but medically plausible
interval-based temporal summaries using semantic
change-type flips only.

CRITICAL REQUIREMENTS:
- You MUST generate exactly ONE incorrect summary
  for EVERY abnormality requested.
- Use ONLY semantic flips (e.g., increase/decrease,
  resolve/persistent).
- Do NOT change the temporal order of intervals.
- Keep laterality unchanged.
- Do NOT accidentally reproduce the correct
  summary.
\end{Verbatim}
\vspace{-1em}
\caption{\textbf{Incorrect global summary prompt.}}
\label{fig:prompt_incorrect_global}
\vspace{-1em}
\end{figure}

\subsection{Generated QA Pair Validation}
\label{sec:dataset_eval}

\subsubsection{Consistency Verification}
\label{sec:appendix_verification}

We apply LLM-assisted verification to ICR, ICR Variant, and GTS questions, where answer options include LLM-generated summaries or distractors.
Temporal Event Localization (TEL) is excluded because TEL questions are constructed deterministically from expert-annotated presence transitions:
both correct and incorrect options are fully specified by rules without free-form generation.
Therefore, LLM-based verification would add little value for TEL and may introduce unnecessary noise~\cite{ZHANG2026129493}.

\paragraph{Objective and Non-circularity}
The goal of verification is \emph{not} to assess clinical correctness or approximate expert judgment.
All clinically meaningful semantics (presence/absence and change types) are inherited directly from expert-validated radiology annotations.
GPT-5.1~\citep{openai2025gpt51} is used strictly as a \emph{consistency checker} to detect violations of predefined logical/semantic constraints in textual options.
Crucially, the model is never asked to determine labels, choose the correct answer, or revise annotations.
All correctness labels are deterministically defined prior to LLM invocation, and the verification output does not affect label assignment.

\paragraph{Verification Granularity}
Verification is performed at the answer-option level, checking that
(i) the correct option faithfully summarizes the expert annotations, and
(ii) each incorrect option is clearly inconsistent with those annotations while remaining syntactically well-formed and medically plausible.

\paragraph{Criteria for Option Validation}
Each generated option is evaluated according to its intended type:
\begin{itemize}[leftmargin=1.3em, labelsep=0.5em, topsep=0.25ex, itemsep=0.25ex, parsep=0pt]
    \item \textbf{Correct Summary Option}
    Must match the expert-annotated presence states and temporal change directions,
    without adding new findings, omitting any relevant interval, or drifting semantically from the annotations.

    \item \textbf{Incorrect (Error) Option}
    Must contradict the annotated change pattern while remaining medically plausible.
    It must not partially match the ground truth or admit an alternative interpretation consistent with the annotations.

    \item \textbf{Ambiguous Option (Rejected)}
    Options using vague or state-based descriptors (e.g., \textit{stable}, \textit{persistent}, \textit{almost resolved})
    are rejected because they do not permit a definitive consistency judgment at the interval level.
\end{itemize}

\paragraph{Ambiguous Change Taxonomy and Refinement}
Most verification failures stem from change descriptors that are ill-posed for interval-level temporal reasoning,
as they encode persistent states, gradual trends, or qualitative impressions without a clear temporal boundary.
To ensure each question admits a single, well-defined interpretation, we remove questions derived from such ambiguous categories
and regenerate questions using only change taxonomies with explicit, directional semantics.

\begin{table}[H]
\centering
\small
\setlength{\tabcolsep}{4pt}
\begin{tabular}{@{}lcc@{}}
\toprule
\textbf{Type} & \textbf{Correct (\%)} & \textbf{Incorrect (\%)} \\
\midrule
ICR          & 99.1 & 97.6 \\
ICR Variant  & 99.6 & 98.4 \\
GTS          & 98.8 & 96.9 \\
\bottomrule
\end{tabular}
\caption{\textbf{Results of LLM-assisted consistency verification after taxonomy refinement.}}
\label{tab:verification_results}
\vspace{-1em}
\end{table}

As shown in Table~\ref{tab:verification_results}, high consistency rates indicate that taxonomy refinement effectively removes semantically unstable cases while preserving the underlying task distribution.
Importantly, this step does not simplify the visual reasoning required; it only prevents ambiguity that can either artificially deflate performance
or reward inconsistent reasoning strategies.
Overall, LLM-assisted verification provides a scalable and reproducible quality-control mechanism that complements expert-validated annotations
without introducing new clinical judgments.

\subsection{Statistics}
\label{sec:stat_break}

\subsubsection{Dataset Statistics}

We present here the detailed dataset statistics, including the distribution of answer choices, the composition of question types, and the list of abnormalities covered (Tables~\ref{tab:answer_distribution}--\ref{tab:abnormality_list}).

\begin{table}[H]
\centering
\small
\begin{tabular}{c r r}
\toprule
\textbf{Answer} & \textbf{\# Questions} & \textbf{Ratio (\%)} \\
\midrule
A & 1062 & 20.00 \\
B & 1006 & 18.94 \\
C & 1109 & 20.88 \\
D & 1209 & 22.76 \\
E & 925  & 17.42 \\
\midrule
\textbf{Total} & \textbf{5311} & \textbf{100.00} \\
\bottomrule
\end{tabular}
\caption{\textbf{Distribution of answer choices across the dataset.}}
\label{tab:answer_distribution}
\vspace{-1em}
\end{table}

\begin{table}[H]
\centering
\small
\begin{tabular}{l r}
\toprule
\textbf{Question Type} & \textbf{\# Questions} \\
\midrule
Single-entity interval summary      & 1000 \\
Multi-entity interval summary       & 1000 \\
Single resolution                   & 500  \\
Resolution-to-emergence             & 500  \\
Single emergence                    & 500  \\
Emergence-to-resolution             & 500  \\
ICR               & 317  \\
Multiple emergence (type 2)          & 250  \\
Multiple emergence (type 1)          & 250  \\
Multiple resolution (type 1)         & 250  \\
Multiple resolution (type 2)         & 244  \\
\midrule
\textbf{Total}                       & \textbf{5311} \\
\bottomrule
\end{tabular}
\caption{\textbf{Distribution of question types across the dataset.}}
\label{tab:qtype_distribution}
\end{table}
\vspace{-1em}

\begin{table}[H]
\centering
\small
\begin{tabular}{l r}
\toprule
\textbf{Abnormality} & \textbf{Count} \\
\midrule
Pleural effusion            & 1672 \\
Pneumothorax                & 996  \\
Cardiomegaly                & 922  \\
Pulmonary edema             & 773  \\
Sub-diaphragmatic air       & 725  \\
Consolidation               & 341  \\
Vascular congestion         & 322  \\
Elevated hemidiaphragm      & 313  \\
Lung opacity                & 313  \\
Atelectasis                 & 311  \\
Bony structures intact      & 296  \\
Normal cardiac silhouette   & 272  \\
\bottomrule
\end{tabular}
\caption{\textbf{List of abnormalities used in question construction and their frequencies.}}
\label{tab:abnormality_list}
\end{table}

\subsubsection{Temporal Pattern Distribution}

We first clarify the definition of abnormality presence states used in our temporal pattern analysis.
The binary states \texttt{pos} (present) and \texttt{neg} (absent) are derived exclusively from
expert-annotated radiology labels provided in the source taxonomy.
Only abnormalities that are explicitly annotated as either present or absent by board-certified
radiologists are considered.
Annotations marked as uncertain, equivocal, or implicitly inferred are excluded from this analysis.

As a result, each five-visit window is represented as a certainty sequence reflecting
definitive abnormality presence or absence at each visit.
This ensures that all temporal state transitions used for TEL construction
(e.g., \texttt{neg} $\rightarrow$ \texttt{pos} for emergence and
\texttt{pos} $\rightarrow$ \texttt{neg} for resolution)
are grounded in high-confidence, clinician-verified annotations rather than heuristic or model-derived signals.

We analyze the temporal state patterns of abnormality presence within five-visit windows
prior to final question filtering, focusing specifically on Temporal Event Localization (TEL).
Each window is represented as a binary certainty sequence indicating abnormality absence
(\texttt{neg}) or presence (\texttt{pos}) across visits.

\begin{table}[H]
\centering
\small
\begin{tabular}{p{5.6cm} r}
\toprule
\textbf{Window Certainty Pattern} & \textbf{Count} \\
\midrule
neg-neg-neg-neg-neg (persistent absence) & 759,512 \\
pos-pos-pos-pos-pos (persistent presence) & 155,056 \\
neg-neg-neg-neg-pos (single emergence) & 21,184 \\
pos-pos-pos-pos-neg (single resolution) & 19,312 \\
Mixed / non-monotonic patterns & 213,304 \\
\bottomrule
\end{tabular}
\caption{\textbf{Distribution of abnormality state patterns in five-visit windows before TEL filtering.}}
\label{tab:tel_window_patterns}
\end{table}

As shown in Table~\ref{tab:tel_window_patterns}, the majority of candidate windows exhibit trivial patterns with no temporal events, such as persistent absence
(\texttt{neg-neg-neg-neg-neg}) or persistent presence (\texttt{pos-pos-pos-pos-pos}).
Without additional filtering, these patterns would dominate TEL questions and reduce the task to identifying the absence of any meaningful temporal change.

Event-based TEL questions are derived from specific state transitions within a window,
where an emergence corresponds to a transition from \texttt{neg} to \texttt{pos},
and a resolution corresponds to a transition from \texttt{pos} to \texttt{neg}.
Such transitions are comparatively rare in the raw distribution.
We therefore apply balancing and filtering strategies to ensure that TEL questions
contain a diverse and representative set of emergence and resolution events,
rather than being dominated by trivial no-event cases.

\subsubsection{Study Date Interval Statistics}
\begin{figure}[h]
\centering
\includegraphics[width=0.98\linewidth]{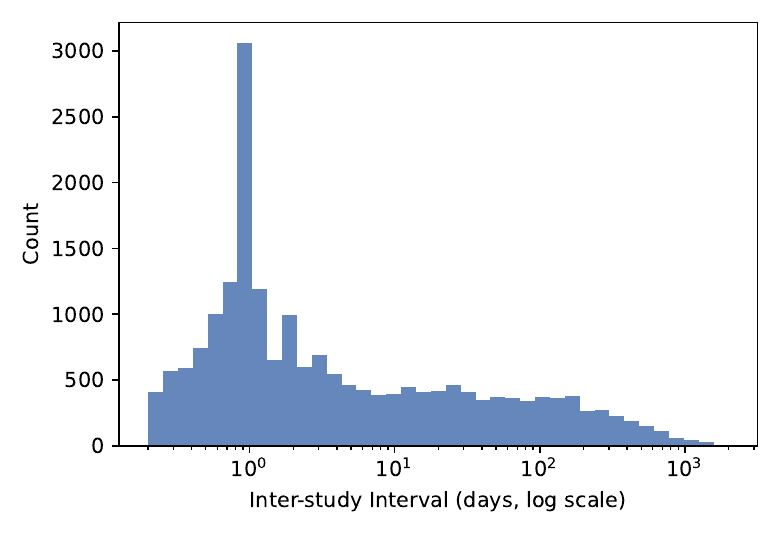}
\vspace{-1em}
\caption{\textbf{Aggregated distribution of inter-study intervals across all five-study windows.} While the median gap is on the order of one to two days, a substantial fraction of intervals spans several months or longer, indicating heterogeneous temporal horizons within the dataset.}
\label{fig:interstudy_gap_distribution}
\vspace{-1.3em}
\end{figure}

To characterize the temporal horizon covered by each five-study window,
we analyze the distribution of study date intervals across the constructed dataset.
For each sample, we consider both interval-level gaps between consecutive studies
(T1$\rightarrow$T2 through T4$\rightarrow$T5) and the total temporal span
from the first to the last study (T1$\rightarrow$T5).
All statistics are reported in days.
Figure~\ref{fig:interstudy_gap_distribution} visualizes the aggregated
distribution of inter-study intervals on a logarithmic scale, highlighting the pronounced heavy-tailed nature of temporal spacing despite short median gaps.

\label{sec:study_interval_stats}

\paragraph{Interval-level Gaps}
Across all consecutive study pairs, the median inter-study gap ranges from
1.4 to 1.9 days, indicating that many follow-up examinations occur within
a short time frame.
However, the mean gaps are substantially larger (36--48 days),
reflecting a pronounced heavy-tailed distribution.
As shown in Table~\ref{tab:interstudy_interval_stats},
the 90th percentile of interval gaps exceeds 95 days for all transitions,
and the maximum gap spans more than four years in some cases.
This demonstrates that, despite short median intervals,
a non-negligible fraction of samples involves long-term follow-up.

\begin{table}[H]
\centering
\scriptsize
\begin{tabular}{lcccccc}
\toprule
\textbf{Interval}
& \textbf{Median}
& \textbf{Mean}
& \textbf{P25}
& \textbf{P75}
& \textbf{P90}
& \textbf{Max} \\
\midrule
T1 $\rightarrow$ T2 & 1.58 & 48.07 & 0.71 & 17.71 & 133.69 & 1609.16 \\
T2 $\rightarrow$ T3 & 1.39 & 41.15 & 0.72 & 14.70 & 113.27 & 1605.58 \\
T3 $\rightarrow$ T4 & 1.55 & 36.13 & 0.76 & 14.78 & 96.89  & 1404.88 \\
T4 $\rightarrow$ T5 & 1.89 & 41.54 & 0.81 & 18.06 & 111.91 & 1506.09 \\
\bottomrule
\end{tabular}
\caption{
\textbf{Summary statistics of inter-study gaps (in days) between consecutive studies within five-study windows.}}
\label{tab:interstudy_interval_stats}
\vspace{-1em}
\end{table}

\paragraph{Total Temporal Span}
We further examine the total temporal span covered by each five-study window by summing the four consecutive inter-study gaps (Table~\ref{tab:window_temporal_span}).
The median window span is approximately 22.5 days, while the interquartile range extends from 4.0 to 193.5 days.
Notably, the 90th percentile exceeds 580 days, and the maximum span approaches five years.
This wide range indicates that a fixed number of visits does not imply a fixed temporal reasoning horizon.

\begin{table}[H]
\centering
\small
\begin{tabular}{lc}
\toprule
\textbf{Statistic} & \textbf{Value (days)} \\
\midrule
Median & 22.45 \\
Mean   & 166.89 \\
P25    & 4.04 \\
P75    & 193.49 \\
P90    & 580.15 \\
Max    & 1708.87 \\
\bottomrule
\end{tabular}
\caption{\textbf{Summary statistics of the total temporal span of five-study windows (T1$\rightarrow$T5).}}
\label{tab:window_temporal_span}
\vspace{-1em}
\end{table}

\paragraph{Interval Bin Distribution}
To provide an interpretable summary of temporal spacing,
we additionally report the distribution of inter-study gaps using
coarse-grained time bins with respect to the number of day (Table~\ref{tab:interstudy_interval_bins}).
Approximately one-third of all consecutive studies occur within one day,
and an additional 30--32\% fall within one week.
At the same time, more than 20\% of intervals exceed 30 days,
and 5--7\% extend beyond six months.
These results confirm that short-term and long-term follow-ups coexist
within the same benchmark.

\begin{table}[H]
\centering
\small
\begin{tabular}{lccccc}
\toprule
\textbf{Interval}
& \textbf{$<$1}
& \textbf{1--7}
& \textbf{7--30}
& \textbf{30--180}
& \textbf{$>$180} \\
\midrule
T1 $\rightarrow$ T2 & 37.4 & 30.9 & 11.0 & 13.2 & 7.5 \\
T2 $\rightarrow$ T3 & 38.2 & 31.0 & 11.9 & 12.3 & 6.7 \\
T3 $\rightarrow$ T4 & 36.8 & 32.2 & 12.2 & 13.1 & 5.7 \\
T4 $\rightarrow$ T5 & 34.4 & 31.9 & 13.3 & 13.9 & 6.5 \\
\bottomrule
\end{tabular}
\caption{\textbf{Distribution of inter-study intervals across five-study windows, reported as percentages.}}
\label{tab:interstudy_interval_bins}
\end{table}

\paragraph{Implications for Longitudinal Reasoning}
Taken together, these statistics indicate that the proposed benchmark
requires models to reason over heterogeneous temporal scales,
even under a fixed five-study window.
Models must handle both subtle short-term changes occurring over days
and substantial longitudinal evolution spanning months or years.
This heterogeneity contributes to the difficulty of Temporal Event Localization,
Interval-wise Change Reasoning, and Global Trajectory Summarization tasks,
and distinguishes our dataset from prior settings that assume
uniform or narrowly constrained time intervals.

\FloatBarrier
\section{Question Templates}
\label{sec:question_templates}

We present representative examples of the question templates used in our benchmark for each task family. The purpose of these examples is not to provide exhaustive coverage, but to illustrate how longitudinal image sequences are paired with structured multiple-choice questions that require temporal reasoning beyond local or pairwise comparisons.

For data privacy and ethical considerations, the image sequences and answer options shown here do not correspond to actual patient timelines included in the released benchmark.
All images are independently selected and assembled solely for illustrative purposes, and no semantic, temporal, or clinical correspondence should be assumed between the displayed image sequences and the textual answer choices.

The actual benchmark instances are constructed exclusively from curated datasets under approved usage terms and are distributed in anonymized form.

\subsection{Temporal Event Localization (TEL)}

\subsubsection{Singe and Multi E/R}
\begin{figure}[H]
\centering
\includegraphics[width=0.9\linewidth]{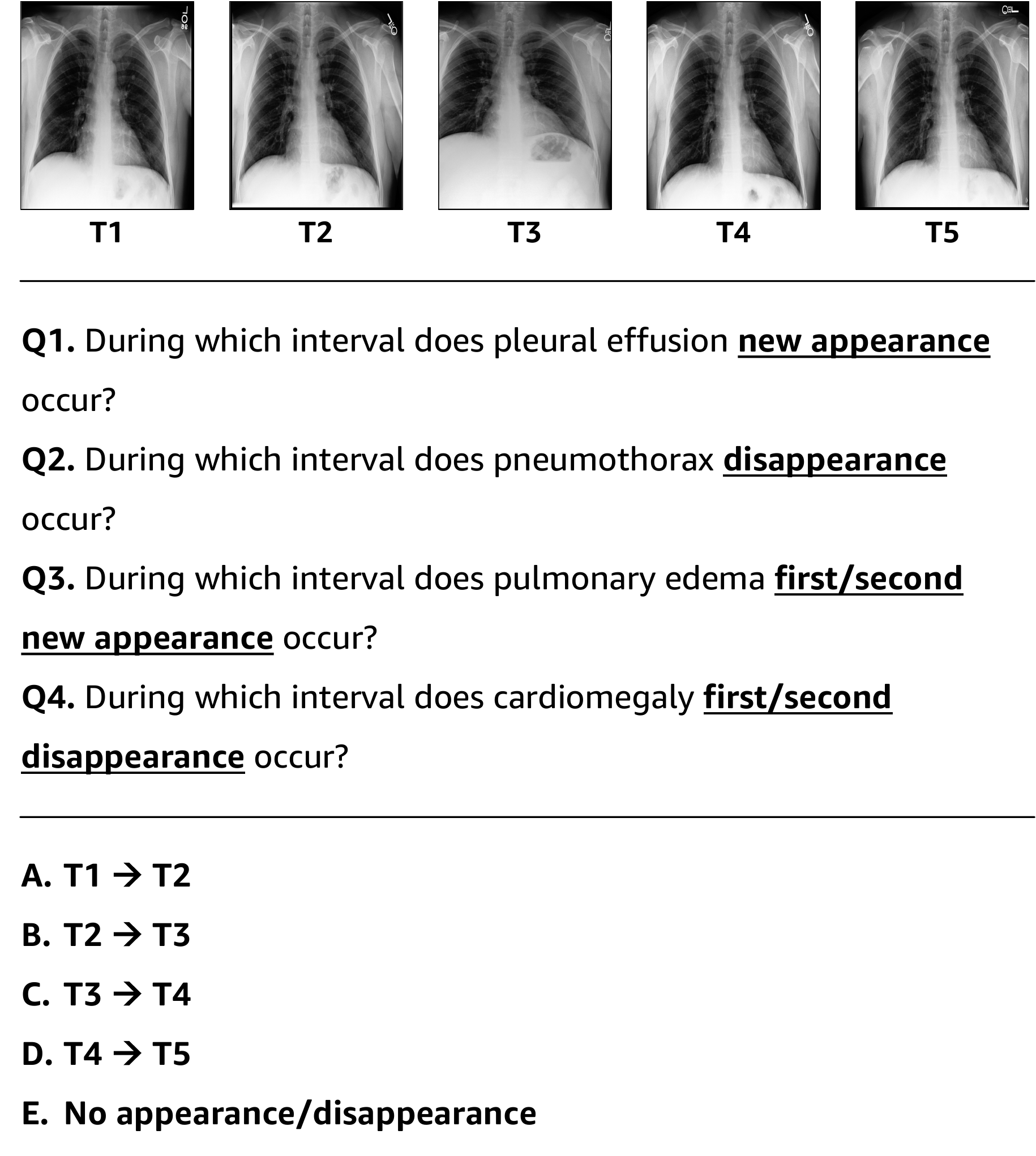} 
\caption{\textbf{Example of Temporal Event Localization (TEL) question} with the Single Emergence (Q1), Single Resolution (Q2), Multi Emergence (Q3), and Multi Resolution (Q4).}
\label{fig:tel_er}
\vspace{-1em}
\end{figure}

Figure~\ref{fig:tel_er} illustrates a Temporal Event Localization (TEL) question with the single/multiple clinically meaningful event. Given a fixed five-visit timeline (T1–T5), the model must identify the unique interval during which an abnormality newly appears or resolves. Correctly answering the Single E/R question requires scanning the entire timeline and comparing all adjacent intervals, rather than relying on a single salient image or a predefined image pair.
Also, the model must distinguish between multiple instances of the same event type (e.g., first versus second emergence) and select the correct interval accordingly in Multi E/R question. This formulation enforces exclusivity among multiple valid-looking temporal candidates, requiring the model to distinguish between first and subsequent events of the same type.

\subsubsection{E→R and R→E}
\begin{figure}[H]
\centering
\includegraphics[width=0.9\linewidth]{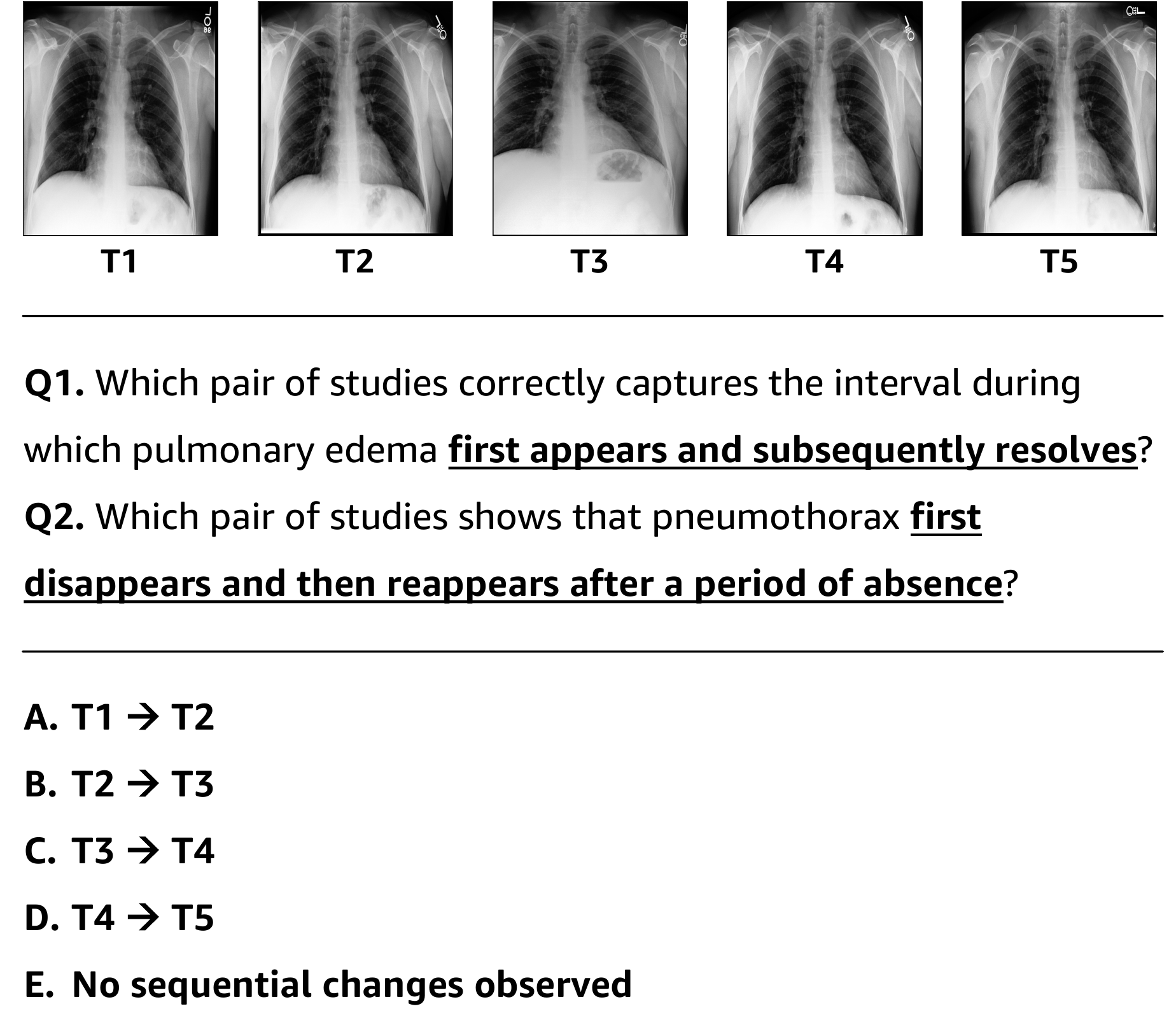}
\caption{\textbf{Example of Temporal Event Localization (TEL) question} with E→R (Q1) and R→E (Q2).}
\label{fig:tel_er_re}
\end{figure}

Figure~\ref{fig:tel_er_re} highlights a TEL question that requires reasoning over ordered event patterns, such as emergence followed by resolution or vice versa.
Rather than identifying a single event in isolation, the model must correctly bind two temporally ordered events across the timeline.
This question type explicitly tests whether models can compose local observations into structured temporal sequences.
Unlike single-event localization, this question requires composing two temporally ordered events into a coherent pattern (e.g., emergence followed by resolution). The task explicitly tests whether models can bind independently recognized events into a structured temporal sequence, rather than detecting them in isolation.

\subsection{Interval-wise Change Reasoning (ICR)}

\begin{figure}[H]
\centering
\includegraphics[width=0.9\linewidth]{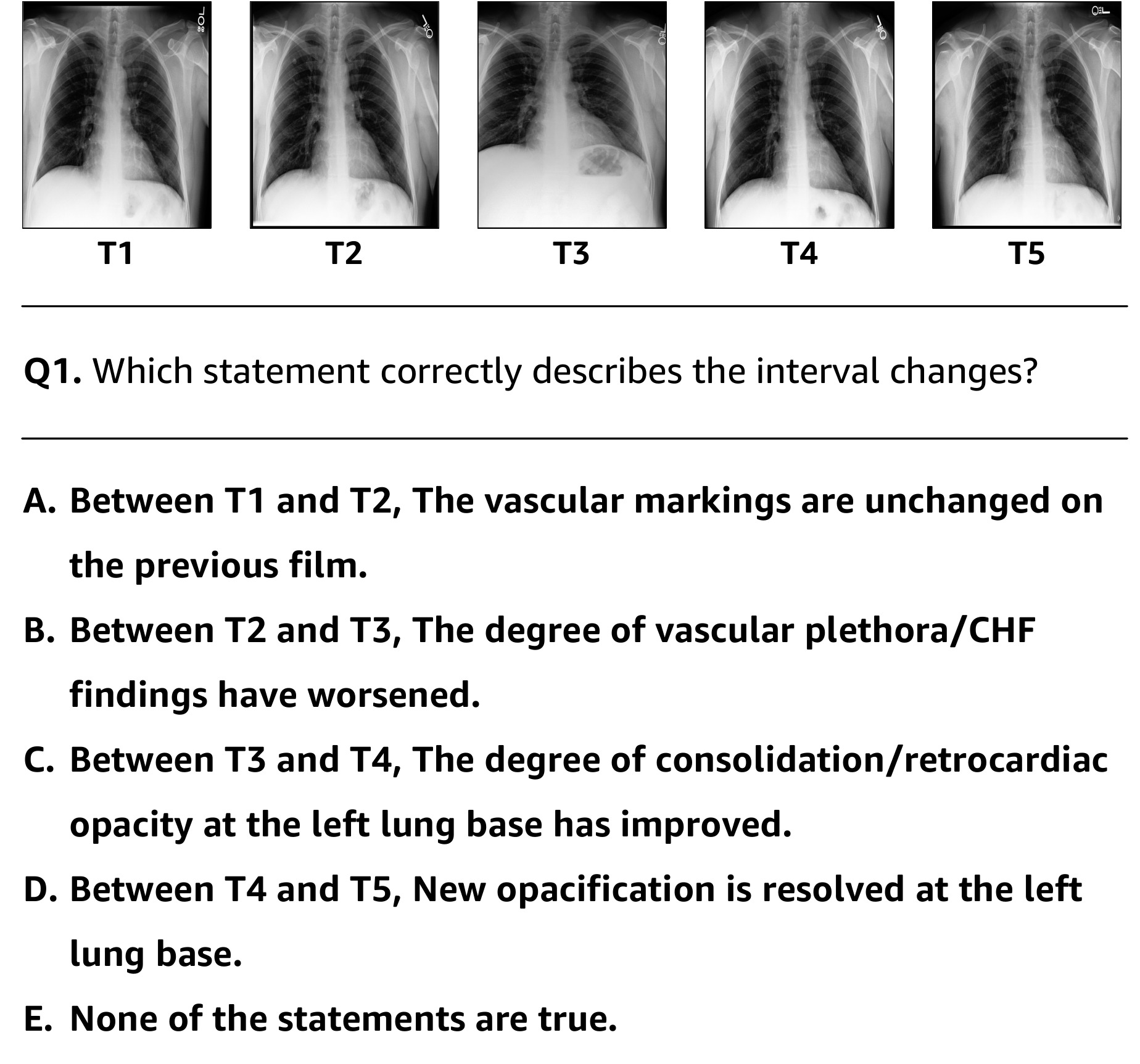} 
\caption{\textbf{Example of Interval-wise Change Reasoning (ICR) question.}}
\label{fig:icr}
\end{figure}

Figure~\ref{fig:icr} shows an Interval-wise Change Reasoning (ICR) question.
The model is presented with a five-visit timeline and must select the statement that correctly describes the visual change occurring within a specific interval.
Unlike pairwise comparison tasks, the relevant interval is not pre-specified, requiring the model to first identify which interval is being described before evaluating the change itself. This design decouples interval selection from change interpretation, making the task sensitive to errors in temporal grounding rather than visual perception alone.

\subsection{Global Trajectory Summarization (GTS)}

\subsubsection{Single Abnormality}
\begin{figure}[H]
\centering
\includegraphics[width=0.9\linewidth]{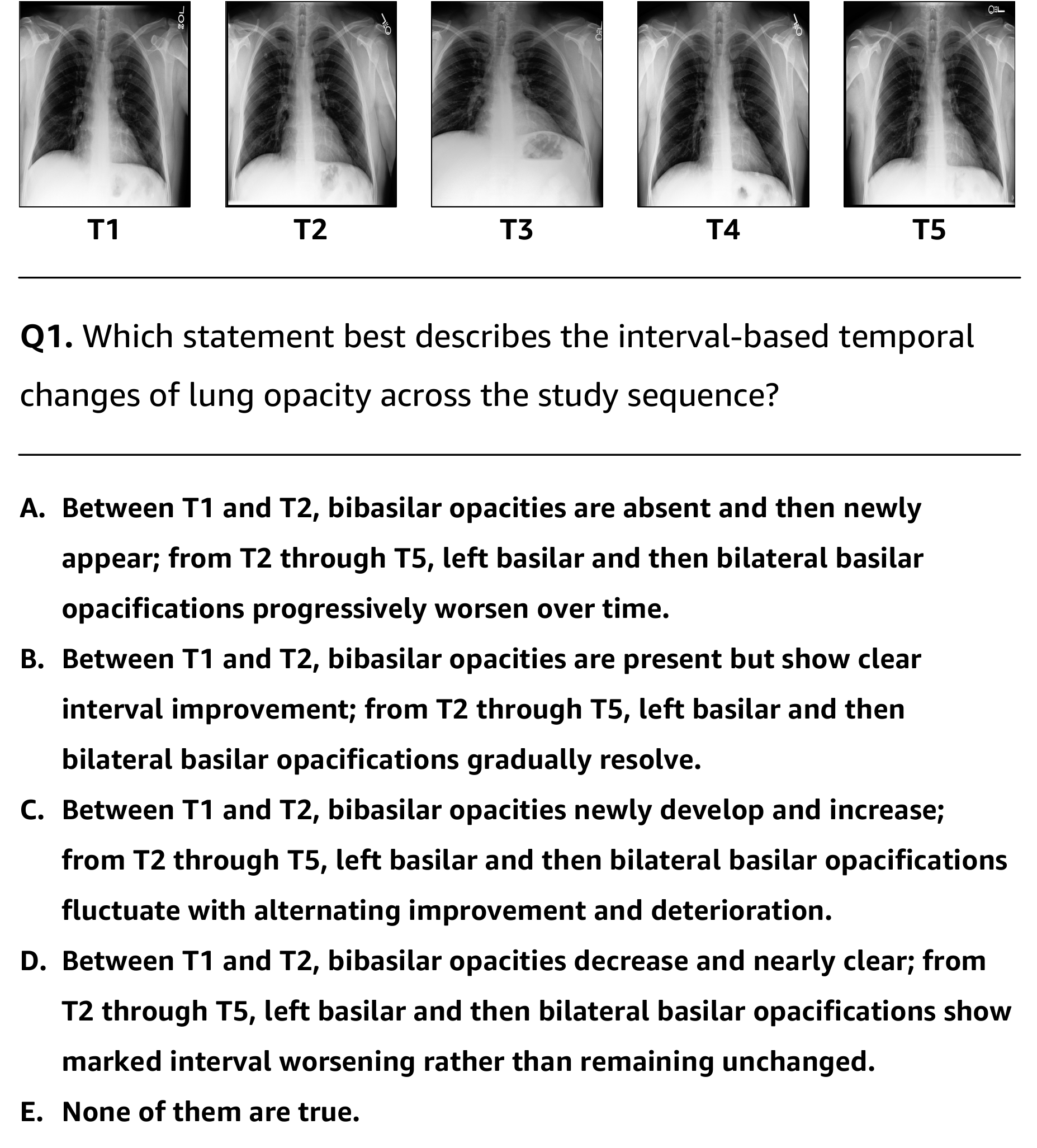} 
\caption{\textbf{Example of Global Trajectory Summarization (GTS) question for a single abnormality.}}
\label{fig:gts_single}
\end{figure} 

Figure~\ref{fig:gts_single} illustrates the Global Trajectory Summarization (GTS) question for a single abnormality.
The model must integrate interval-level changes across the entire timeline and select the option that best characterizes the overall disease course.
No single interval or image is sufficient to answer this question. Correct reasoning requires aggregating interval-level changes across the entire timeline to infer the overall trajectory of a single abnormality.

\begin{table*}[t!]
\small
\resizebox{\textwidth}{!}{
\begin{tabular}{llccccccc}
\toprule
\multirow{2}{*}{\textbf{Category}}
& \multirow{2}{*}{\textbf{Model}} 
& \multicolumn{3}{c}{\textbf{TEL}} 
& \textbf{ICR} 
& \multicolumn{2}{c}{\textbf{GTS}} 
& \multirow{2}{*}{\textbf{Overall}} \\
\cmidrule(lr){3-5} \cmidrule(lr){6-6} \cmidrule(lr){7-8}
& 
& \textbf{Single (E/R)} 
& \textbf{Multiple (E/R)} 
& \textbf{E$\rightarrow$R / R$\rightarrow$E} 
& \textbf{--} 
&\textbf{Multi Abnormality}
& \textbf{Single Abnormality}
&  \\

\midrule
\multirow{3}{*}{\textbf{Closed}}
& Claude Sonnet 4.5      & 0.217 & 0.226 & 0.234 & 0.454 & 0.292 & 0.391 & 0.316 \\
& Gemini 3.0 Preview    & 0.259 & 0.318 & 0.251 & 0.416 & 0.399 & \textbf{0.547} & 0.370 \\
& GPT-5.2               & \textbf{0.327} & \textbf{0.345} & \textbf{0.333} & 0.426 & 0.385 & 0.545 & 0.396 \\

\midrule
\multirow{7}{*}{\textbf{General}}
& InternVL3.5-8B        & 0.233 & 0.262 & 0.186 & 0.543 & 0.362 & 0.377 & 0.346 \\
& InternVL3.5-14B       & 0.245 & 0.267 & 0.246 & 0.325 & 0.268 & 0.378 & 0.292 \\
& InternVL3.5-38B       & 0.289 & 0.282 & 0.222 & \textbf{0.546} & \textbf{0.486} & 0.498 & \textbf{0.401} \\
& QwenVL3-8B            & 0.228 & 0.281 & 0.198 & 0.164 & 0.237 & 0.317 & 0.231 \\
& QwenVL3-32B           & 0.253 & 0.259 & 0.233 & 0.227 & 0.326 & 0.370 & 0.273 \\
& DeepSeek-VL-16B       & 0.193 & 0.128 & 0.199 & 0.180 & 0.187 & 0.158 & 0.175 \\
& IDEFICS2-8B           & 0.147 & 0.283 & 0.274 & 0.243 & 0.179 & 0.235 & 0.229 \\

\midrule
\multirow{4}{*}{\textbf{Medical}}
& Lingshu-7B            & 0.219 & 0.261 & 0.170 & 0.180 & 0.180 & 0.319 & 0.218 \\
& Lingshu-32B           & 0.206 & 0.236 & 0.207 & 0.221 & 0.285 & 0.355 & 0.249 \\
& MedGemma-4B           & 0.186 & 0.245 & 0.315 & 0.287 & 0.198 & 0.264 & 0.253 \\
& MedGemma-27B          & 0.214 & 0.328 & 0.245 & 0.420 & 0.215 & 0.262 & 0.293 \\

\bottomrule
\end{tabular}
}
\captionof{table}{
\textbf{Performance of evaluated models with decoding temperature set to 0.7.}
Results are reported across task families and question subtypes.
Compared to the default low-temperature setting, higher temperature generally leads
to degraded or unstable performance on TEL questions, while effects on ICR and GTS
are more model-dependent.
}
\label{tab:temp07_results}
\end{table*}

\subsubsection{Multi Abnormality}
\begin{figure}[H]
\centering
\includegraphics[width=0.9\linewidth]{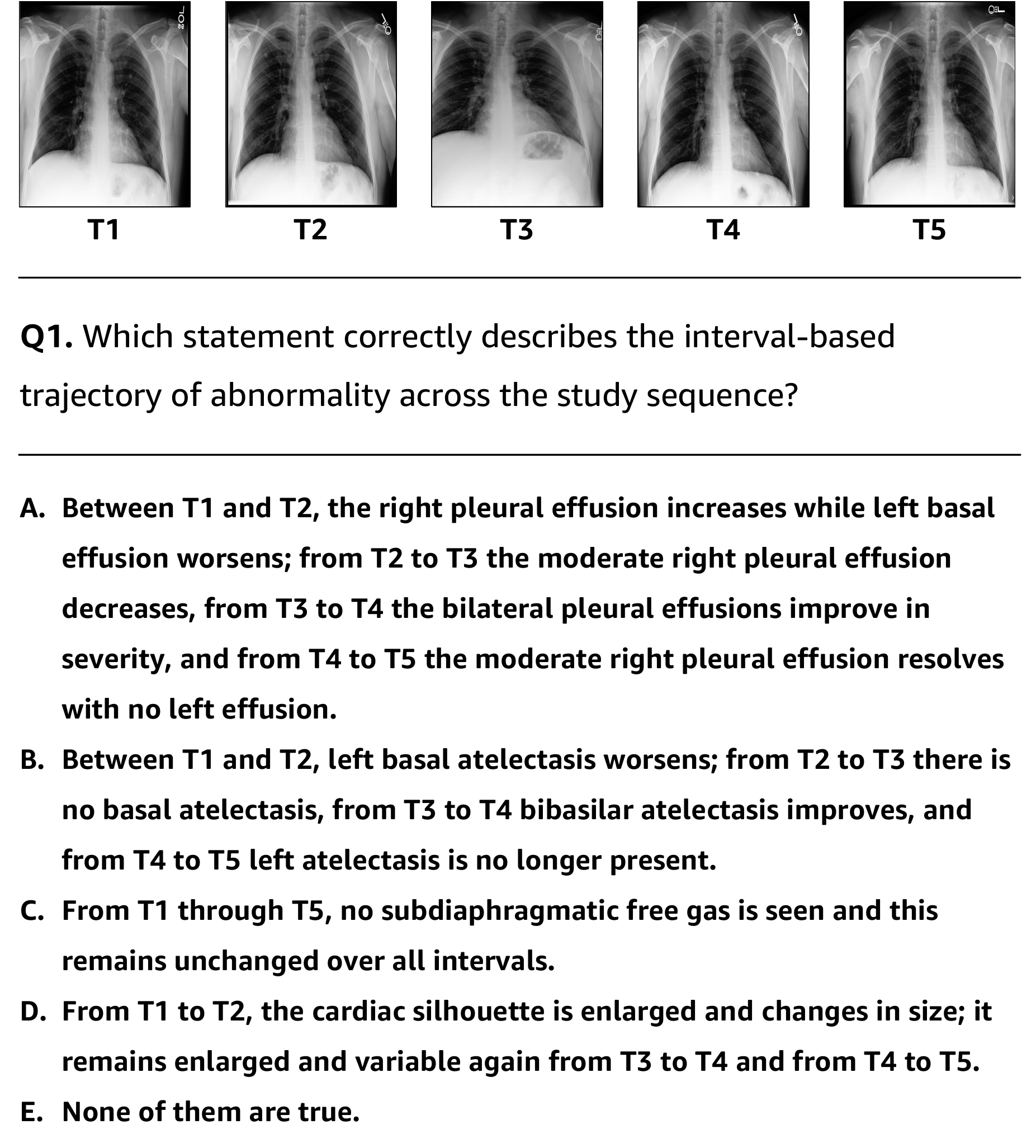} 
\caption{\textbf{Example of Global Trajectory Summarization (GTS) question for multiple abnormality.}}
\label{fig:gts_multi}
\end{figure} 

Figure~\ref{fig:gts_multi} presents a GTS question involving multiple abnormalities.
Each answer option describes a different abnormality trajectory, and the model must identify the one that correctly summarizes the longitudinal changes observed in the timeline.
This formulation increases difficulty by requiring both global temporal integration and correct abnormality selection under mutual exclusivity.

\section{Sensitivity to Decoding Temperature}
\label{sec:appendix_temperature}

To examine whether model performance is sensitive to decoding stochasticity, we additionally evaluate all models using a higher decoding temperature of 0.7, which is a commonly adopted default setting in many large language model deployments.
Table~\ref{tab:temp07_results} and Table~\ref{tab:icr_variant_category} summarizes the results across all task families under this setting.

\begin{table}[H]
\centering
\small
\begin{tabular}{llc}
\toprule
\textbf{Category} & \textbf{Model} & \textbf{ICR Variant} \\
\midrule
\multirow{3}{*}{\textbf{Closed}}
& Claude Sonnet 4.5      & 0.607 \\
& Gemini 3.0 Preview    & 0.705 \\
& GPT-5.2               & \textbf{0.776} \\
\midrule
\multirow{7}{*}{\textbf{General}}
& InternVL3.5-8B        & 0.667 \\
& InternVL3.5-14B       & 0.623 \\
& InternVL3.5-38B       & 0.683 \\
& QwenVL3-8B            & 0.590 \\
& QwenVL3-32B           & 0.612 \\
& DeepSeek-VL-16B       & 0.246 \\
& IDEFICS2-8B           & 0.421 \\
\midrule
\multirow{4}{*}{\textbf{Medical}}
& Lingshu-7B            & 0.574 \\
& Lingshu-32B           & 0.612 \\
& MedGemma-4B           & 0.612 \\
& MedGemma-27B          & 0.694 \\
\bottomrule
\end{tabular}
\caption{\textbf{Performance comparison on the ICR Variant across model categories with decoding temperature set to 0.7.}
All questions involve single-abnormality interval-level change reasoning with the interval explicitly specified.}
\label{tab:icr_variant_category}
\end{table}

Overall, we observe that increasing the decoding temperature does not fundamentally improve performance on the proposed benchmark.
In particular, Temporal Event Localization (TEL) performance consistently degrades or becomes more unstable across models, reflecting the increased susceptibility of precise temporal localization to stochastic generation.
For Interval-wise Change Reasoning (ICR) and Global Trajectory Summarization (GTS), the effects of higher temperature are model-dependent and do not yield systematic gains.

These results indicate that the challenges posed by our benchmark are not attributable to overly restrictive decoding settings.
Instead, even under a commonly used higher-temperature regime, models continue to struggle with long-horizon temporal reasoning over longitudinal medical image sequences.

\subsection{Performance Stability}
To assess result reliability, we report mean accuracy, standard deviation, and 95\% confidence intervals for representative models across three independent inference runs under the main evaluation setting (temperature = 0.7).

\begin{table}[h!]
\centering
\small
\begin{tabular}{lcccc}
\toprule
\textbf{Model} & \textbf{Mean} & \textbf{Std} & \textbf{95\% CI} \\
\midrule
GPT-5.2         & 0.401 & 0.009 & [0.379, 0.423] \\
InternVL3.5-38B & 0.391 & 0.004 & [0.383, 0.400] \\
MedGemma-27B    & 0.276 & 0.006 & [0.261, 0.291] \\
\bottomrule
\end{tabular}
\caption{\textbf{Performance stability of representative models.}}
\label{tab:performance_stability}
\end{table}

As shown in Table~\ref{tab:performance_stability}, 
all three representative models exhibit narrow 
standard deviations and tight confidence intervals, 
confirming that the reported results are stable 
across inference runs. The low variance observed 
across model categories—closed-source (GPT-5.2), 
open-source (InternVL3.5-38B), and 
medical-specialized (MedGemma-27B)—indicates that 
performance differences between models reflect 
genuine capability gaps rather than run-to-run 
stochasticity. 

Taken together with the 
temperature sensitivity analysis in the preceding 
section, these results establish that the 
consistently low accuracy observed on MI-CXR is 
robust to evaluation conditions and not an artifact 
of decoding randomness.

\section{ICR Variant}
\label{sec:icr_variant}
\begin{figure}[H]
\centering
\includegraphics[width=0.98\linewidth]{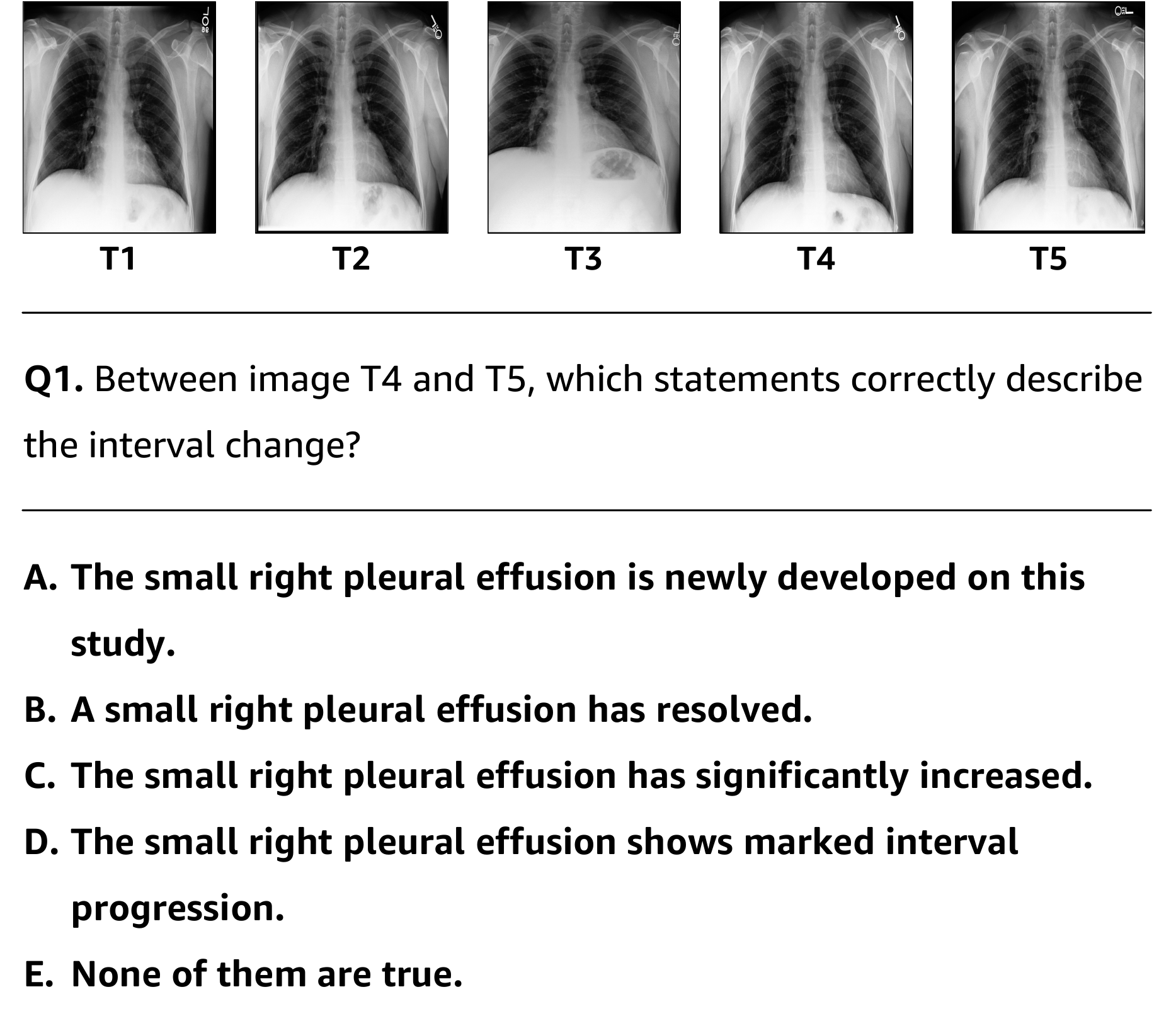}
\caption{\textbf{Example of Interval-wise Change Reasoning (ICR) variant.}}
\label{fig:icr_var}
\end{figure}

We introduce an Interval-wise Change Reasoning (ICR) variant to provide a more controlled evaluation setting.
Unlike the original ICR task, which requires models to both identify the relevant temporal interval and interpret the corresponding change, this variant explicitly specifies the interval of interest (see Figure~\ref{fig:icr_var}).

Each question presents a fixed five-visit timeline and asks the model to determine which statement correctly describes the visual change occurring within a given interval (e.g., T4 $\rightarrow$ T5).
All questions in this variant focus on a single abnormality and assess only change-type interpretation, such as new appearance, resolution, or progression.
Interval localization and multi-abnormality reasoning are intentionally excluded.

The evaluation set consists of 400 questions constructed under this setting.
By isolating interval-level change interpretation, this variant reduces ambiguity arising from temporal grounding and enables more direct assessment of a model’s ability to recognize and characterize visual changes across longitudinal medical images.


\subsection{Prompt Template for ICR Variant Question Generation}
\label{sec:icr_variant_prompt}

To generate distractor answer choices for the ICR variant, we employ GPT-5.1~\citep{openai2025gpt51} under strict constraints.
The model is used solely for surface-level language generation and does not determine correctness or temporal labels.
All change types are pre-defined based on expert annotations.

The prompt (Figure~\ref{fig:icr_var_prompt}) enforces the following constraints:
\begin{itemize}[leftmargin=1.3em, labelsep=0.5em, topsep=0.25ex, itemsep=0.25ex, parsep=0pt]
    \item the same abnormality and anatomical region must be preserved,
    \item only interval-level change semantics may be modified,
    \item no new findings, organs, or laterality changes are allowed, and
    \item ambiguous modifiers are explicitly prohibited.
\end{itemize}

\begin{figure}[h!]
\centering
\small
\begin{Verbatim}[fontsize=\scriptsize, frame=single]
You generate distractor statements for a radiology interval-
change question.

INPUT:
- Correct interval-change statement.

RULES:
- Preserve the same abnormality and anatomical region.
- Describe interval-level change only.
- Generate statements that are clearly false relative to the
  correct meaning.
- Do not introduce new abnormalities, organs, or laterality.
- Avoid ambiguous expressions (e.g., "slightly worsened").

ALLOWED CATEGORIES:
- Resolution
- New Appearance
- Marked Improvement
- Severe Worsening

OUTPUT:
Return exactly N distractor sentences.
One sentence per line. No numbering or formatting.
\end{Verbatim}
\caption{\textbf{Prompt for ICR variant generation.}}
\label{fig:icr_var_prompt}
\end{figure}

All final answer correctness labels are assigned deterministically prior to LLM invocation.

\section{Stage-wise Evaluation Protocol and Implementation Details}
\label{sec:appendix_eval_protocol}

This section describes the evaluation protocol and implementation details used to assess model performance across all tasks.
The goal of this section is to clarify how models are evaluated in a consistent and reproducible manner, rather than to analyze performance differences or model limitations.

\subsection{Overview of the Evaluation Pipeline}

All evaluated models follow a unified evaluation pipeline. Each model is provided with an identical sequence of longitudinal chest X-ray images
and a task-specific question formulated in a standardized format. Model outputs are processed using deterministic rules to extract discrete answer choices, which are then compared against ground-truth labels.

The evaluation pipeline consists of three main steps:
\begin{itemize}[leftmargin=1.3em, labelsep=0.5em, topsep=0.25ex, itemsep=0.25ex, parsep=0pt]
    \item preparation of model inputs according to task-specific guidelines,
    \item model inference following a structured, stage-wise procedure, and
    \item rule-based answer extraction and scoring.
\end{itemize}

This design ensures that differences in performance reflect model capability
rather than variations in evaluation methodology.

\subsection{Stage-wise Inference Procedure}

To encourage explicit temporal reasoning, we adopt a stage-wise inference procedure
for all tasks.
In the first stage, models are prompted to generate intermediate descriptions
that focus on interval-level visual changes observed across the image sequence.
These intermediate outputs are constrained by task-specific guidelines to prevent
premature answer selection or reliance on global shortcuts.

In the second stage, models are instructed to select a discrete answer option
based solely on the intermediate representations produced in the first stage.
This separation between perception-oriented reasoning and answer selection
helps ensure that models explicitly process temporal information before committing
to a final decision.

Importantly, this stage-wise structure is applied uniformly across all models,
with no task-specific tuning or model-dependent adjustments during evaluation.

The stage-wise inference design is motivated by the observation that
end-to-end answer prediction often encourages shortcut reasoning,
where models directly map visual cues to answer options without
explicitly reasoning over temporal structure.
By separating interval-level description from answer selection,
the evaluation protocol encourages models to externalize their
temporal reasoning process~\cite{lee2025evaluatingstepbystepreasoningtraces, wei2023chainofthoughtpromptingelicitsreasoning, wang2025mrgagentsmultiagentframeworkimproved, jiang2025comtchainofmedicalthoughtreduceshallucination, smit2020chexbertcombiningautomaticlabelers, Guo__2024, kyung2025predictingtemporalchangespatients}.

This design does not constrain model capacity or expressiveness,
but instead enforces a reasoning order that mirrors how longitudinal
medical images are interpreted in practice.

\subsection{Role of Evaluation Guidelines}

Task-specific evaluation guidelines are designed to constrain model behavior
without encoding task-specific heuristics.
Rather than prescribing how an answer should be derived,
the guidelines prevent degenerate strategies such as ignoring intermediate images,
collapsing multi-interval reasoning into a single comparison,
or exploiting superficial textual cues.

By standardizing reasoning boundaries across tasks,
the guidelines ensure that performance differences reflect
a model’s ability to process temporal information,
rather than its sensitivity to prompt phrasing.

\subsection{Task-specific Evaluation Guidelines}

While the overall evaluation pipeline is shared across tasks,
each task employs distinct guidelines that reflect its underlying reasoning requirements.

Across all tasks, the evaluation protocol enforces interval-level reasoning
as a common intermediate step.
TEL focuses on identifying the precise temporal location of an event,
ICR evaluates the interpretation of changes within a specific interval,
and GTS assesses the integration of multiple interval-level observations
into a coherent global trajectory.

Despite these differences, all tasks share a unified evaluation philosophy:
models are required to reason explicitly over temporal structure,
and answers are scored deterministically without human intervention.

\subsubsection{Temporal Event Localization (TEL)}

TEL questions assess a model’s ability to identify the temporal interval
during which a specific abnormality emerges or resolves.
All TEL questions are constructed deterministically from expert-annotated
presence transitions and do not rely on free-form textual summaries.

During evaluation, models are guided to reason explicitly about changes
between consecutive image pairs.
Answer selection is based directly on identifying the interval
that satisfies the queried event condition.
As TEL questions do not involve LLM-generated summaries or distractor construction,
they are evaluated using fully deterministic rules.

\subsubsection{Interval-wise Change Reasoning (ICR)}

ICR questions require models to interpret the visual change occurring
within a specific temporal interval.
Evaluation guidelines instruct models to focus on interval-level descriptions
rather than global trends, ensuring that reasoning remains grounded
in the designated time span.

For the ICR variant, the target interval is explicitly specified,
further isolating change interpretation from interval localization.
In both cases, models are evaluated based on their ability to correctly
map interval-level observations to the appropriate answer option.

\subsubsection{Global Trajectory Summarization (GTS)}

GTS questions evaluate a model’s ability to integrate information
across multiple intervals and reason about the overall temporal trajectory
of an abnormality.
Although GTS requires global reasoning, models are still guided to first
consider interval-level changes before producing a summary judgment.

Evaluation guidelines emphasize consistency across the entire timeline,
penalizing answers that rely on isolated observations
or ignore intermediate temporal patterns.

\subsection{Answer Extraction and Scoring}

Model outputs are evaluated using a deterministic scoring procedure.
For each response, a single answer option is extracted using
rule-based parsing of the model output.
Responses that contain multiple answer choices, ambiguous selections,
or unparseable formats are treated as incorrect.

No partial credit or fuzzy matching is applied.
All tasks use exact-match scoring against predefined ground-truth answers,
ensuring consistency and fairness across models and question types.

Deterministic scoring eliminates ambiguity in evaluation
and ensures that reported results are fully reproducible.
By avoiding partial credit or subjective judgment,
the evaluation protocol provides a strict but transparent assessment
of temporal reasoning performance.

All evaluation procedures, guidelines, and scoring rules
are applied uniformly across all evaluated models.
No model-specific adaptations or post-hoc adjustments are introduced,
ensuring that comparisons reflect intrinsic model capability
rather than evaluation artifacts.

\section{Qualitative Error Analysis}
\label{sec:additional_examples}

\subsection{Failure Mode in TEL}
\label{sec:failure_tel_abstention}

\vspace{1em}

\begin{tcolorbox}[breakable, colback=white, colframe=black, boxrule=0.4pt, left=6pt,right=6pt,top=6pt,bottom=6pt]
\small
\textbf{Problem (TEL -- Single Emergence):}

\raggedright
During which interval does \underline{pneumothorax newly} \underline{appear}?

\vspace{0.4em}

\textbf{Answer Choices:}
\begin{itemize}[leftmargin=1.4em, labelsep=0.5em, topsep=0.25ex, itemsep=0.25ex, parsep=0pt]
    \item[A.] T1 $\rightarrow$ T2
    \item[B.] T2 $\rightarrow$ T3
    \item[C.] T3 $\rightarrow$ T4
    \item[D.] T4 $\rightarrow$ T5
    \item[E.] No appearance
\end{itemize}

\vspace{0.6em}

\textbf{Stage-1 Model Response (Interval-level Description):}
\begin{itemize}[leftmargin=1.3em, labelsep=0.5em, topsep=0.25ex, itemsep=0.25ex, parsep=0pt]
    \item \textbf{T1--T2:} No visible pneumothorax is observed. [\cmark]
    \item \textbf{T2--T3: A new pneumothorax becomes apparent.} [\cmark]
    \item \textbf{T3--T4:} The pneumothorax persists with similar extent. [\cmark]
    \item \textbf{T4--T5:} The pneumothorax resolves. [\cmark]
\end{itemize}

\vspace{0.6em}

\textbf{Model Prediction:} \textcolor{red}{C (T3 $\rightarrow$ T4)} [\xmark] \\
\textbf{Ground Truth:} B (T2 $\rightarrow$ T3) [\cmark]

\vspace{0.4em}

\textbf{Failure Mechanism:}
The model observes the correct local change but fails to translate it into a consistent global decision.

\end{tcolorbox}
\vspace{-1em}
\captionof{figure}{\textbf{Failures in Temporal Event Localization - Single (E/R).}}
\label{fig:f_tel_single}

\vspace{1em}

\begin{tcolorbox}[breakable, colback=white, colframe=black, boxrule=0.4pt, left=6pt,right=6pt,top=6pt,bottom=6pt]
\small
\textbf{Problem (TEL -- Multiple Emergence Candidates):}

\raggedright
During which interval does \underline{pleural effusion second} \underline{appearance occur}?

\vspace{0.4em}

\textbf{Answer Choices:}
\begin{itemize}[leftmargin=1.4em, labelsep=0.5em, topsep=0.25ex, itemsep=0.25ex, parsep=0pt]
    \item[A.] T1 $\rightarrow$ T2
    \item[B.] T2 $\rightarrow$ T3
    \item[C.] T3 $\rightarrow$ T4
    \item[D.] T4 $\rightarrow$ T5
    \item[E.] No second appearance of pleural effusion
\end{itemize}

\vspace{0.6em}

\textbf{Stage-1 Model Response (Interval-level Description):}
\begin{itemize}[leftmargin=1.3em, labelsep=0.5em, topsep=0.25ex, itemsep=0.25ex, parsep=0pt]
    \item \textbf{T1--T2:} No pleural effusion is observed. [\cmark]
    \item \textbf{T2--T3:} A small pleural effusion becomes visible. [\cmark]
    \item \textbf{T3--T4:} Pleural effusion becomes more conspicuous bilaterally. [\cmark]
    \item \textbf{T4--T5:} Pleural effusion remains present. [\cmark]
\end{itemize}

\vspace{0.6em}

\textbf{Model Prediction:} \textcolor{red}{B (T2 $\rightarrow$ T3)} [\xmark] \\
\textbf{Ground Truth:} E (No second appearance of pleural effusion) [\cmark]

\vspace{0.4em}

\textbf{Failure Mechanism:}
The failure stems from the model’s inability to track ordinal emergence events, treating the first appearance as the only salient temporal change.

\end{tcolorbox}
\vspace{-1em}
\captionof{figure}{\textbf{Failures in Temporal Event Localization - Multiple (E/R).}}
\label{fig:f_tel_multi}

\begin{tcolorbox}[breakable, colback=white, colframe=black, boxrule=0.4pt, left=6pt,right=6pt,top=6pt,bottom=6pt]
\small
\textbf{Problem (TEL -- Multiple Emergence Candidates):}

Which pair of studies correctly captures the interval during which pleural effusion \underline{first appears subsequently} \underline{resolves}?

\vspace{0.4em}

\textbf{Answer Choices:}
\begin{itemize}[leftmargin=1.4em, labelsep=0.5em, topsep=0.25ex, itemsep=0.25ex, parsep=0pt]
    \item[A.] T2, T3
    \item[B.] T1, T3
    \item[C.] T3, T4
    \item[D.] T1, T5
    \item[E.] There is no sequential emergence to resolution
\end{itemize}

\vspace{0.6em}

\textbf{Stage-1 Model Response (Interval-level Description):}
\begin{itemize}[leftmargin=1.3em, labelsep=0.5em, topsep=0.25ex, itemsep=0.25ex, parsep=0pt]
    \item \textbf{T1--T2:} A small pleural effusion is \textcolor{blue}{newly observed at T2}. [\cmark]
    \item \textbf{T2--T3:} A small pleural effusion is \textcolor{blue}{no longer visible}, leading to resolution at T3. [\cmark]
    \item \textbf{T3--T4:} At T4, there is no visual evidence of pleural effusion. [\cmark]
    \item \textbf{T4--T5:} Pleural effusion is not visible with similar extent. [\cmark]
\end{itemize}

\vspace{0.6em}

\textbf{Model Prediction:} \textcolor{red}{C (T3, T4)} [\xmark] \\
\textbf{Ground Truth:} A (T2, T3) [\cmark]

\vspace{0.4em}

\textbf{Failure Mechanism:}
Although the model correctly identifies the emergence and resolution events, it fails assigning temporal decisiveness.

\end{tcolorbox}
\vspace{-1em}
\captionof{figure}{\textbf{Failures in Temporal Event Localization - (E$\rightarrow$R / R$\rightarrow$E).}}
\label{fig:f_tel_erre}

\vspace{-0.5em}

\subsection{Failure Mode in ICR}
\label{sec:failure_icr_forced}

\vspace{0.5em}

\begin{tcolorbox}[breakable, colback=white, colframe=black, boxrule=0.4pt, left=6pt,right=6pt,top=6pt,bottom=6pt]
\small
\textbf{Problem (ICR -- Multiple Abnormalities):}

Which statement \underline{correctly describes} the interval-level change?

\vspace{0.4em}

\textbf{Answer Choices:}
\begin{itemize}[leftmargin=1.4em, labelsep=0.5em, topsep=0.25ex, itemsep=0.25ex, parsep=0pt]
    \item[A.] Between T1 and T2, the pneumothorax has increased.
    \item[B.] Between T2 and T3, the pleural effusion has decreased.
    \item[C.] Between T3 and T4, lung opacity has worsened.
    \item[D.] Between T4 and T5, bibasilar atelectasis has resolved.
    \item[E.] None of the statements are true.
\end{itemize}

\vspace{0.6em}

\textbf{Stage-1 Model Response (Interval-level Description):}
\begin{itemize}[leftmargin=1.3em, labelsep=0.5em, topsep=0.25ex, itemsep=0.25ex, parsep=0pt]
    \item \textbf{T1--T2 (Pneumothorax):} No visible pneumothorax at either timepoint. [\cmark]
    \item \textbf{T2--T3 (Pleural effusion):} A small pleural effusion is present at both timepoints, with \textcolor{blue}{only marginal and ambiguous change}. [\cmark]
    \item \textbf{T3--T4 (Lung opacity):} Patchy lung opacities appear similar in extent, without clear progression. [\cmark]
    \item \textbf{T4--T5 (Atelectasis):} Bibasilar atelectasis remains visible, with no definite interval resolution. [\cmark]
\end{itemize}

\vspace{0.6em}

\textbf{Model Prediction:} \textcolor{red}{B (Between T2 and T3, the pleural effusion has decreased)} [\xmark] \\
\textbf{Ground Truth:} E (None of the statements are true) [\cmark]

\vspace{0.4em}

\textbf{Failure Mechanism:}
Although all interval-level observations are plausible but ambiguous,
the model commits to a specific change direction without sufficient evidence,
leading to incorrect decision.

\end{tcolorbox}
\vspace{-1em}
\captionof{figure}{\textbf{Failures in Interval-wise Change Reasoning.}}
\label{fig:f_icr}

\subsection{Failure Mode in GTS}
\label{sec:gts_failure}

\vspace{1em}

\begin{tcolorbox}[breakable, colback=white, colframe=black, boxrule=0.4pt, left=6pt,right=6pt,top=6pt,bottom=6pt]
\small
\textbf{Problem (GTS -- Single Abnormality):}

Which statement \underline{best describes} the interval-based temporal changes of pleural effusion
across the study sequence?

\vspace{0.4em}

\textbf{Answer Choices:}
\begin{itemize}[leftmargin=1.4em, labelsep=0.5em, topsep=0.25ex, itemsep=0.25ex, parsep=0pt]
    \item[A.] Right pleural effusion newly appears, worsens, and a new left effusion develops.
    \item[B.] Right pleural effusion develops over time with a worsening left effusion.
    \item[C.] Right pleural effusion resolves completely with a worsening left effusion.
    \item[D.] Right pleural effusion progressively increases with worsening left effusion.
    \item[E.] None of them are true.
\end{itemize}

\vspace{0.6em}

\textbf{Stage-1 Model Response (Interval-level Description):}
\begin{itemize}[leftmargin=1.3em, labelsep=0.5em, topsep=0.25ex, itemsep=0.25ex, parsep=0pt]
    \item \textbf{T1--T2:} Right pleural effusion appears similar; small left effusion persists. [\cmark]
    \item \textbf{T2--T3:} Right pleural effusion decreases; left remains similar. [\cmark]
    \item \textbf{T3--T4:} \textcolor{blue}{Right pleural effusion resolved}; left slightly increases. [\cmark]
    \item \textbf{T4--T5:} No evidence of right pleural effusion with \textcolor{blue}{left pleural effusions markedly worsen}. [\cmark]
\end{itemize}

\vspace{0.6em}

\textbf{Model Prediction:} \textcolor{red}{D} [\xmark] \\
\textbf{Ground Truth:} C [\cmark]

\vspace{0.4em}
\textbf{Failure Mechanism:}
The model correctly identifies all local interval changes but fails to integrate them into a coherent global trajectory, instead imposing an incorrect monotonic worsening pattern.

\end{tcolorbox}
\vspace{-1em}
\captionof{figure}{\textbf{Failures in Global Trajectory Summarization -- Single Abnormality.}}
\label{fig:f_gts_single}

\vspace{1em}

\begin{tcolorbox}[breakable, colback=white, colframe=black, boxrule=0.4pt, left=6pt,right=6pt,top=6pt,bottom=6pt]
\small
\textbf{Problem (GTS -- Multi Abnormality):}

Which statement \underline{correctly describes} the interval-based trajectory of abnormalities across the study sequence?

\vspace{0.4em}

\textbf{Target Abnormalities:}
Pleural effusion, pneumothorax, lung opacity, bony structures intact

\vspace{0.4em}

\textbf{Answer Choices:}
\begin{itemize}[leftmargin=1.4em, labelsep=0.5em, topsep=0.25ex, itemsep=0.25ex, parsep=0pt]
    \item[A.] A pneumothorax newly appears between T2 and T3 and resolves by T5, while pleural effusions remain absent.
    \item[B.] Pleural effusions progressively worsen bilaterally across all intervals, with steadily worsening lung opacities.
    \item[C.] No pneumothorax is seen; bony structures remain intact; pleural effusion alternates between right-sided presence and absence, and lung opacity fluctuates with overall improvement by T5.
    \item[D.] No pneumothorax is seen; bony structures remain intact; pleural effusion shifts from right-sided to left-sided over time, while lung opacity improves then worsens again.
    \item[E.] None of them are true.
\end{itemize}

\vspace{0.6em}

\textbf{Stage-1 Model Response (Interval-level Description):}
\begin{itemize}[leftmargin=1.3em, labelsep=0.5em, topsep=0.25ex, itemsep=0.25ex, parsep=0pt]
    \item \textbf{T1--T2:} No pneumothorax is observed; bony structures remain intact. A small \textcolor{blue}{right pleural effusion is present}, and left-sided lung opacity worsens. [\cmark]
    \item \textbf{T2--T3:} No pneumothorax is observed; bony structures remain intact. The \textcolor{blue}{right pleural effusion is no longer seen}, and lung opacity improves. [\cmark]
    \item \textbf{T3--T4:} No pneumothorax is observed; bony structures remain intact. A small \textbf{left} pleural effusion newly appears, and lung opacity worsens again. [\cmark]
    \item \textbf{T4--T5:} No pneumothorax is observed; bony structures remain intact. The left pleural effusion persists, and lung opacity worsening persists. [\cmark]
\end{itemize}

\vspace{0.6em}

\textbf{Model Prediction:} \textcolor{red}{E} [\xmark] \\
\textbf{Ground Truth:} D [\cmark]

\vspace{0.4em}
\textbf{Failure Mechanism:}
The model correctly tracks each abnormality locally but fails to jointly integrate them into a single globally consistent multi-entity trajectory, defaulting to a null option.

\end{tcolorbox}
\vspace{-1em}
\captionof{figure}{\textbf{Failures in Global Trajectory Summarization -- Multi Abnormality.}}
\label{fig:f_gts_multi}

\vspace{1em}

Figures~\ref{fig:f_tel_single}--\ref{fig:f_gts_multi} demonstrate that failures in longitudinal reasoning do not primarily stem from isolated visual misperceptions, but rather from systematic breakdowns in organizing and integrating temporal evidence.
The failures arise from systematic difficulties in committing to a single temporal event, enforcing exclusivity among competing hypotheses, calibrating uncertainty under weak evidence, and composing multiple local observations into a global narrative.
By aligning failure modes with task-specific reasoning requirements, our analysis provides clearer insight into the limitations of current VLMs on long-horizon medical image understanding.

\paragraph{Local Interval Misinterpretation}
We note that some errors originate from incorrect interpretation of individual image pairs, such as misidentifying the presence or direction of change within a single interval.
These failures reflect limitations in pairwise visual comparison and are not specific to longitudinal reasoning across multiple timepoints.
As such, they are not the focus of our qualitative analysis, which instead emphasizes reasoning failures that arise even when local interval descriptions are correct.

\begin{tcolorbox}[breakable, colback=white, colframe=black, boxrule=0.4pt, left=6pt,right=6pt,top=6pt,bottom=6pt]
\small

\textbf{Problem (ICR -- Multiple Abnormalities):}

Which statement correctly describes the interval changes?

\vspace{0.4em}

\textbf{Answer Choices:}
\begin{itemize}[leftmargin=1.4em, labelsep=0.5em, topsep=0.25ex, itemsep=0.25ex, parsep=0pt]
    \item[A.] Between T1 and T2, the pneumothorax has increased.
    \item[B.] Between T2 and T3, the cardiac and mediastinal contours have worsened.
    \item[C.] Between T3 and T4, the pleural effusion is stable.
    \item[D.] Between T4 and T5, there has been a resolution in bibasilar atelectasis.
    \item[E.] None of the statements are true.
\end{itemize}

\vspace{0.6em}

\textbf{Stage-1 Model Response (Interval-level Description):}
\begin{itemize}[leftmargin=1.3em, labelsep=0.5em, topsep=0.25ex, itemsep=0.25ex, parsep=0pt]
    \item \textbf{A (T1--T2):} No evidence of pneumothorax shown. [\cmark]
    \item \textbf{B (T2--T3):} Cardiac and mediastinal contours appear similar to the previous study. [\cmark]
    \item \textbf{C (T3--T4):} \textcolor{red}{The pleural effusion seems to be stable compared to the previous study.} [\xmark]
    \item \textbf{D (T4--T5):} Bibasilar atelectasis shows interval resolution. [\cmark]
\end{itemize}

\vspace{0.6em}

\textbf{Model Prediction:} \textcolor{red}{C} [\xmark] \\
\textbf{Ground Truth:} D [\cmark]

\vspace{0.4em}
\textbf{Failure Mechanism:}
A single local misinterpretation (stability of pleural effusion) is directly mapped to an answer choice.

\end{tcolorbox}
\vspace{-1em}
\captionof{figure}{\textbf{Failure within local-interval misinterpretation — ICR example.}}
\label{fig:error_type_C_icr}

\vspace{1em}

\begin{tcolorbox}[breakable, colback=white, colframe=black, boxrule=0.4pt, left=6pt,right=6pt,top=6pt,bottom=6pt]
\small

\textbf{Problem (GTS -- Multi Abnormality):}

Which statement correctly describes the interval-based trajectory of abnormalities
across the study sequence?

\vspace{0.4em}

\textbf{Answer Choices:}
\begin{itemize}[leftmargin=1.4em, labelsep=0.5em, topsep=0.25ex, itemsep=0.25ex, parsep=0pt]
    \item[A.] A small right pleural effusion newly appears between T2 and T3,
    persists through T3 to T4, and resolves by T5, while lung opacities improve
    after an initial worsening and bony structures remain intact.

    \item[B.] Lung opacities progressively worsen across all intervals,
    accompanied by the development of bilateral pleural effusions.

    \item[C.] No pleural effusion is present at any time point, while lung
    opacities fluctuate mildly without a clear trajectory.

    \item[D.] A pneumothorax newly appears between T3 and T4 and resolves by T5,
    with otherwise stable lung findings.

    \item[E.] None of them are true.
\end{itemize}

\vspace{0.4em}

\textbf{Stage-1 Model Response (Interval-level Description):}
\begin{itemize}[leftmargin=1.3em, labelsep=0.5em, topsep=0.25ex, itemsep=0.25ex, parsep=0pt]
    \item \textbf{T1--T2:} No pneumothorax; bony structures intact; no pleural effusion;
    lung opacity worsens. [\cmark]

    \item \textbf{T2--T3:} No pneumothorax; bony structures intact;
    \textcolor{red}{no pleural effusion reported};
    lung opacity improves. [\xmark]

    \item \textbf{T3--T4:} No pneumothorax; bony structures intact;
    \textcolor{red}{pleural effusion described as absent}; lung opacity similar.
    [\xmark]

    \item \textbf{T4--T5:} No pneumothorax; bony structures intact;
    \textcolor{red}{pleural effusion described as absent}; lung opacity stable.
    [\xmark]
\end{itemize}

\vspace{0.4em}

\textbf{Model Prediction:} \textcolor{red}{E (None of them are true)} [\xmark] \\
\textbf{Ground Truth:} A [\cmark]

\vspace{0.4em}

\textbf{Failure Mechanism:}
A minor early omission at T2--T3 eliminates the emergence cue for pleural effusion,
which then propagates to later intervals and shifts the final decision toward an option
that fits lung opacity changes but contradicts the true effusion trajectory.

\end{tcolorbox}
\vspace{-1em}
\captionof{figure}{\textbf{Failure within local-interval misinterpretation — GTS example.}}
\label{fig:error_cascade_gts_multi}

\vspace{1em}

\setlength{\parindent}{1em}
\section{Error Type Distribution Across Tasks}
\label{sec:error_distribution}

This section presents a quantitative analysis of how task-aligned reasoning failures are distributed across task families and model families.
The analysis complements the qualitative examples in Appendix~\ref{sec:additional_examples} by demonstrating that the observed error types arise systematically as a function of task structure and reasoning demands, rather than from isolated perceptual mistakes. 

Importantly, this analysis excludes local interval misinterpretation, which we treat as a baseline source of perceptual noise rather than a task-aligned reasoning failure. By focusing exclusively on higher-level temporal decision errors—such as temporal commitment, exclusivity enforcement, and global evidence integration—we isolate systematic reasoning breakdowns that persist even when interval-level observations are locally plausible or partially correct.

\subsection{Distribution by Task Family}

\begin{figure}[h!]
\centering
\includegraphics[width=\linewidth]{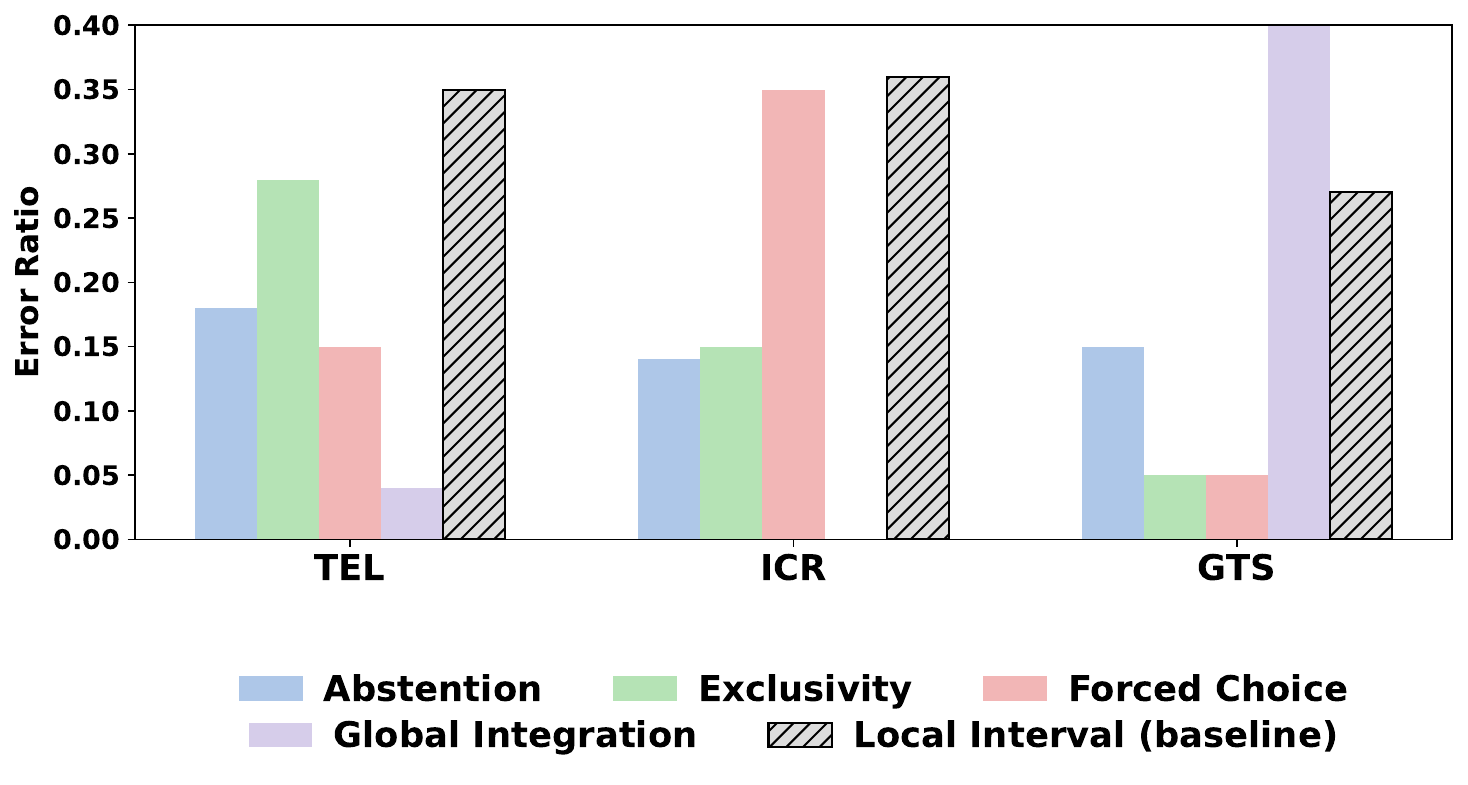} 
\caption{\textbf{Distribution of task-aligned reasoning failure types across TEL, ICR, and GTS.} Local interval misinterpretation is excluded to highlight higher-level temporal reasoning failures.}
\label{fig:error_by_task}
\end{figure}

Figure~\ref{fig:error_by_task} illustrates how different reasoning failures manifest across the three task families.

For Temporal Event Localization (TEL), errors are dominated by Abstention under Uncertainty and Failure of Temporal Exclusivity. In particular, models often fail to enforce temporal exclusivity constraints, such as selecting exactly one decisive onset or resolution interval when multiple candidates appear locally plausible.
This indicates that, although models frequently produce reasonable interval-level descriptions, they struggle to commit to a single decisive interval or to enforce ordinal and exclusivity constraints when multiple candidate intervals appear plausible.
Explicit forced-choice errors are comparatively rare in TEL, suggesting a preference for conservative abstention over over-commitment under temporal ambiguity.

In contrast, Interval-wise Change Reasoning (ICR) exhibits a markedly different failure profile.
Here, Forced Choice under Insufficient Evidence constitutes the dominant error type.
Because ICR questions require selecting a single correct interval-level statement among multiple competing alternatives, even marginal or ambiguous evidence can lead models to over-commit to a specific abnormality or direction of change.
This reflects a systematic difficulty in managing uncertainty when commitment is required at the interval level.

For Global Trajectory Summarization (GTS), errors are overwhelmingly driven by Failures in Global Temporal Integration.
Even when interval-level observations are locally consistent, models often fail to compose these observations into a coherent and globally consistent trajectory across the full study sequence. This highlights that long-horizon temporal aggregation and consistency maintenance remain primary bottlenecks, with even minor interval-level uncertainties or ambiguities propagating into incorrect global summaries.

Overall, these distributions demonstrate that error patterns are strongly task-dependent, reflecting the distinct temporal reasoning constraints imposed by each task family rather than random or model-specific noise.

\subsection{Distribution by Model Family}
\begin{figure}[h!]
\centering
\includegraphics[width=\linewidth]{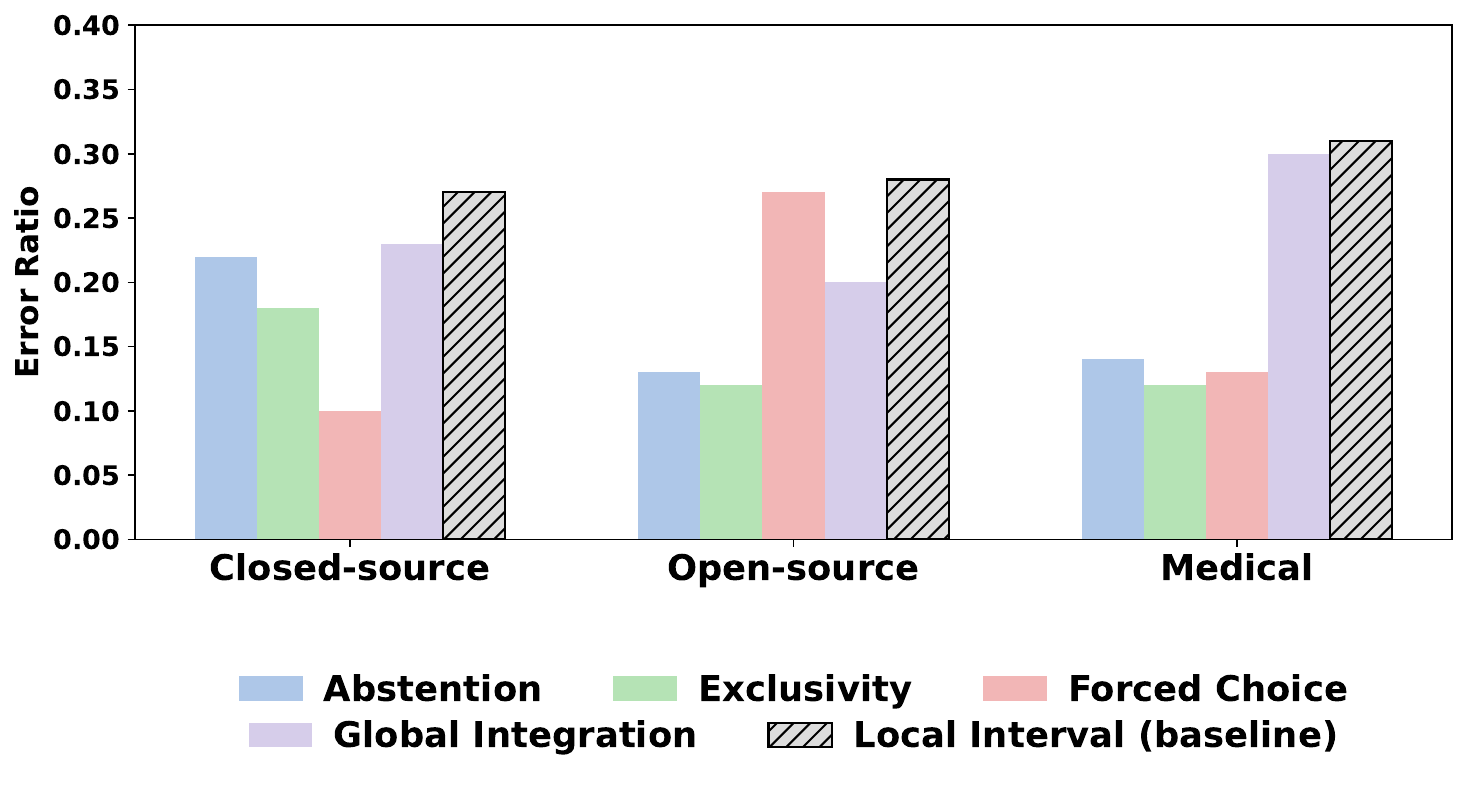} 
\caption{\textbf{Distribution of task-aligned reasoning failures across closed-source, open-source, and medical-specialized VLMs.}}
\label{fig:error_by_model_family}
\end{figure}

Figure~\ref{fig:error_by_model_family} shows how task-aligned reasoning failures differ across model families.

Closed-source models (e.g., GPT-5.2, Claude, Gemini) exhibit a high proportion of Abstention under Uncertainty, reflecting a conservative decision-making strategy.
While this behavior reduces hallucinated or over-confident errors, it also leads to missed correct answers in cases where interval-level reasoning is largely correct but final temporal commitment fails.

Open-source VLMs display a contrasting pattern, with substantially higher rates of Forced Choice under Insufficient Evidence.
These models are more likely to commit to a specific answer even when temporal evidence is weak or ambiguous, resulting in confident but incorrect predictions.
This suggests weaker uncertainty calibration during temporal decision-making.

Medical-specialized VLMs show relatively balanced error profiles across abstention, forced choice, and global integration failures.
Despite their domain-specific training, these models continue to exhibit substantial difficulty in composing temporally distributed evidence into consistent longitudinal interpretations, particularly for GTS-style questions.

These results indicate that model specialization influences how models fail, but does not eliminate higher-level temporal reasoning breakdowns. Notably, these differences reflect distinct temporal decision strategies rather than differences in visual perception, reinforcing that higher-level reasoning failures are strongly task- and structure-dependent.

\subsubsection{Distribution across Representative Models}
\begin{figure}[h!]
\centering
\includegraphics[width=\linewidth]{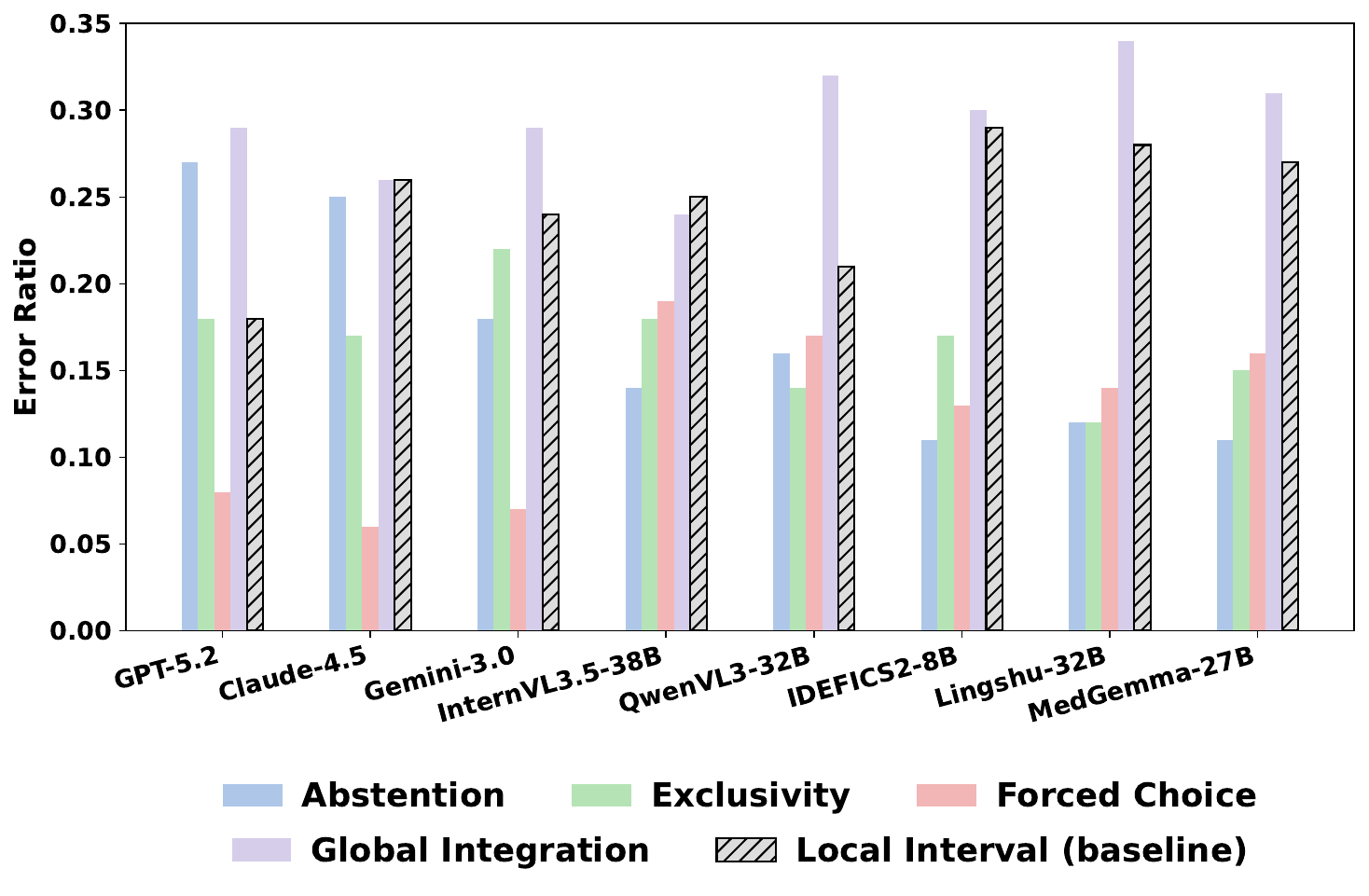} 
\caption{\textbf{Error type distribution} across representative models, illustrating variability in temporal decision-making strategies.}
\label{fig:error_by_representative_models}
\end{figure}

Figure~\ref{fig:error_by_representative_models} further decomposes task-aligned reasoning failures across representative models within each family.

Among closed-source models, GPT-5.2 exhibits the strongest abstention tendency, whereas Gemini shows a relatively higher proportion of exclusivity-related failures, reflecting subtle differences in temporal commitment strategies.
Open-source models exhibit greater heterogeneity: for example, InternVL3.5-38B shows a more balanced distribution across error types, while QwenVL3-32B and IDEFICS2-8B display elevated forced-choice errors.
Medical-specialized models such as Lingshu-32B and MedGemma-27B continue to demonstrate substantial global integration failures, reinforcing that domain specialization alone does not resolve long-horizon temporal reasoning challenges.

\section{Report Generation Pilot Study}
\label{sec:report_generation}

\subsection{Experimental Setup}
\label{sec:report_generation_setup}

We conduct a pilot report generation experiment to assess whether the
temporal reasoning limitations observed in the multiple-choice setting
persist under free-form generation.

\subsection{Input Formulation}
For each sample, we use the same five-visit longitudinal CXR sequence
(T1–T5) and question used in the GTS task. All models are prompted with
a unified instruction designed to elicit interval-based temporal summaries.
Specifically, models are instructed to describe how abnormalities evolve
between consecutive intervals (T1–T2, T2–T3, T3–T4, T4–T5), rather than
providing independent descriptions of each timepoint. The prompt enforces
a concise paragraph format and emphasizes temporal progression across visits.
This ensures consistency across models and isolates temporal reasoning ability
from prompt variation.

\subsection{Reference Construction}
Ground-truth summaries are derived deterministically from the structured
annotations used in the MCQA benchmark. Specifically, we use the correct
answer option from the GTS multiple-choice task as the reference summary,
which is constructed from annotation-grounded temporal changes. This
ensures that evaluation remains fully aligned with the underlying
longitudinal annotations and avoids introducing additional sources of noise.

\subsection{Evaluation Protocol}
We evaluate generated summaries using standard report generation metrics,
including ROUGE-L \cite{rouge}, METEOR \cite{meteor}, and CIDEr \cite{cider}. All models are evaluated under
identical conditions with deterministic decoding where applicable. We use a unified prompt that enforces interval-based temporal reasoning across consecutive visits, as shown in Figure~\ref{fig:report_gen_prompt}.

\begin{figure}[h!]
\centering
\small
\begin{Verbatim}[fontsize=\scriptsize, frame=single]
Below are five chest X-ray images from the same patient,
labeled T1 through T5 in chronological order.

[Question]

Instructions:
1. Focus only on interval-based temporal changes.
2. Describe how the abnormality evolves between consecutive
   intervals (T1–T2, T2–T3, T3–T4, T4–T5).
3. Do not describe each timepoint independently.
4. Provide a concise paragraph.

Example answer formats:
- Between T1 and T2, a left pleural effusion increases; from
  T2 to T3 it remains stable; it worsens from T3 to T4 and
  partially regresses from T4 to T5.
- Bilateral pleural effusions are absent between T1 and T2;
  they newly appear and enlarge from T2 to T3, increase 
  further from T3 to T4, and slightly improve from T4 to T5.

Return only the interval-based summary.
\end{Verbatim}
\caption{\textbf{Prompt for report generation in the pilot study.}
Models are instructed to generate interval-based temporal summaries
over five-visit CXR sequences, emphasizing evolution across
consecutive intervals rather than independent descriptions.}
\label{fig:report_gen_prompt}
\end{figure}

\subsection{Quantitative Results}
\label{sec:report_generation_results}

\begin{table}[h!]
\centering
\small
\begin{tabular}{lccc}
\toprule
\textbf{Model} & \textbf{ROUGE-L} & \textbf{METEOR} & \textbf{CIDEr} \\
\midrule
GPT-5.2 & 0.307 & 0.354 & 0.358 \\
InternVL3.5-38B & 0.341 & 0.369 & 0.411 \\
MedGemma-27B & 0.326 & 0.340 & 0.393 \\
\bottomrule
\end{tabular}
\caption{\textbf{Pilot report generation results on GTS-type longitudinal sequences.} All models exhibit modest performance, indicating challenges in multi-interval temporal integration.}
\label{tab:report_generation_results}
\end{table}

As shown in Table~\ref{tab:report_generation_results}, all evaluated models
achieve relatively modest performance across standard report generation
metrics. While these models are known to perform strongly in single-image
CXR report generation, their performance degrades when required to summarize
multi-interval temporal evolution.

This trend is consistent with our observations in the MCQA setting.
Despite generating locally plausible interval-level descriptions, models
struggle to compose these into globally coherent temporal narratives,
resulting in degraded overall performance. These findings support our
claim that the primary bottleneck lies in longitudinal temporal integration,
rather than the specific output format.

\section{Prompting Analysis}
\label{sec:prompting_analysis}

\subsection{Reasoning-style Prompting}
\label{sec:reasoning_prompting}

\begin{table*}[h!]
\small
\resizebox{\textwidth}{!}{
\begin{tabular}{llccccccc}
\toprule
\multirow{2}{*}{\textbf{Category}}
&
\multirow{2}{*}{\textbf{Model}} 
& \multicolumn{3}{c}{\textbf{TEL}} 
& \textbf{ICR} 
& \multicolumn{2}{c}{\textbf{GTS}} 
& \multirow{2}{*}{\textbf{Overall}} \\
\cmidrule(lr){3-5} \cmidrule(lr){6-6} \cmidrule(lr){7-8}
& 
& \textbf{Single (E/R)} 
& \textbf{Multiple (E/R)} 
& \textbf{E$\rightarrow$R / R$\rightarrow$E} 
& \textbf{--} 
&\textbf{Multi Abnormality}
& \textbf{Single Abnormality}
&  \\

\midrule
\multirow{1}{*}{\textbf{Closed}}
& GPT-5.2               & 0.329 & 0.378 & 0.370 & 0.442 & 0.578 & 0.397 & 0.412 \\

\midrule
\multirow{1}{*}{\textbf{General}}
& InternVL3.5-38B       & 0.251 & 0.304 & 0.312 & 0.495 & 0.546 & 0.453 & 0.381 \\

\midrule
\multirow{1}{*}{\textbf{Medical}}
& MedGemma-27B          & 0.220 & 0.325 & 0.261 & 0.457 & 0.271 & 0.222 & 0.272 \\

\bottomrule
\end{tabular}
}
\captionof{table}{
\textbf{Performance under reasoning-style prompting (“Let’s think step by step”).}
}
\label{tab:reasoning_prompting}
\end{table*}

We evaluate whether reasoning-style guidance improves performance
on longitudinal tasks by adding explicit instructions such as
\textit{“Let’s think step by step”}~\cite{NEURIPS2022_8bb0d291} under the same deterministic
decoding setting used in the main experiments.

As shown in Table~\ref{tab:reasoning_prompting}, reasoning-style prompting
does not yield consistent improvements across tasks. While some models
exhibit marginal gains on specific subtasks, overall performance remains
comparable to or only slightly above the zero-shot baseline.

This suggests that the primary challenge does not lie in eliciting
step-by-step reasoning, but in the model’s ability to reliably integrate
temporal evidence across multiple intervals. Even when encouraged to
produce structured reasoning, models continue to struggle with enforcing
temporal constraints and maintaining global consistency.

\subsection{One-shot Prompting}
\label{sec:fewshot_prompting}

\begin{table*}[h!]
\small
\resizebox{\textwidth}{!}{
\begin{tabular}{llccccccc}
\toprule
\multirow{2}{*}{\textbf{Category}}
&
\multirow{2}{*}{\textbf{Model}} 
& \multicolumn{3}{c}{\textbf{TEL}} 
& \textbf{ICR} 
& \multicolumn{2}{c}{\textbf{GTS}} 
& \multirow{2}{*}{\textbf{Overall}} \\
\cmidrule(lr){3-5} \cmidrule(lr){6-6} \cmidrule(lr){7-8}
& 
& \textbf{Single (E/R)} 
& \textbf{Multiple (E/R)} 
& \textbf{E$\rightarrow$R / R$\rightarrow$E} 
& \textbf{--} 
& \textbf{Multi Abnormality}
& \textbf{Single Abnormality}
&  \\

\midrule
\multirow{1}{*}{\textbf{Closed}}
& GPT-5.2               & 0.322 & 0.360 & 0.350 & 0.475 & 0.531 & 0.385 & 0.394 \\

\midrule
\multirow{1}{*}{\textbf{General}}
& InternVL3.5-38B       & 0.285 & 0.320 & 0.235 & 0.589 & 0.425 & 0.498 & 0.366 \\

\midrule
\multirow{1}{*}{\textbf{Medical}}
& MedGemma-27B          & 0.237 & 0.359 & 0.257 & 0.424 & 0.283 & 0.242 & 0.288 \\

\bottomrule
\end{tabular}
}
\captionof{table}{
\textbf{Performance under 1-shot prompting.}
All models are evaluated under deterministic decoding without additional system-level role specification.
Performance does not consistently improve compared to zero-shot settings and may degrade,
indicating that few-shot demonstrations do not effectively alleviate multi-interval temporal reasoning challenges.
}
\label{tab:fewshot_prompting}
\end{table*}

We further evaluate whether providing a single demonstration example
(1-shot prompting) improves model performance. All experiments are
conducted under deterministic decoding, without introducing additional
system-level role specifications.

As shown in Table~\ref{tab:fewshot_prompting}, 1-shot prompting does not
consistently improve performance and in several cases leads to slight
degradation compared to the zero-shot baseline. This indicates that model errors are unlikely to stem from misunderstanding of task format. Instead, longitudinal reasoning over multi-interval sequences requires case-specific integration of temporally dependent visual evidence, which cannot be substantially alleviated by a single demonstration example. Moreover, introducing demonstration examples increases the effective input
length by adding additional image sequences, which may dilute attention
and contribute to performance instability, particularly for open-source
models.

Overall, these results suggest that the observed performance limitations
reflect intrinsic challenges in multi-interval temporal reasoning rather
than prompt design or task-format ambiguity.












\end{document}